\documentclass[sigconf,9pt,nonacm]{acmart}

\usepackage{amsmath}
\usepackage{graphicx}
\usepackage{booktabs}
\usepackage{csquotes}
\usepackage{pifont}    
\usepackage{algorithm}
\usepackage{algpseudocode}
\usepackage{multirow}
\usepackage{xcolor}     
\usepackage{amsmath}    
\usepackage{tabularx}   
\usepackage{array}      
\usepackage{fontawesome5}
\usepackage{colortbl}
\definecolor{darkgreen}{rgb}{0.0, 0.5, 0.0}
\definecolor{lorablue}{rgb}{0.1, 0.1, 0.6}
\newcolumntype{C}{>{\centering\arraybackslash}X} 

\setcopyright{acmlicensed}
\copyrightyear{2026}
\acmYear{2026}
\acmDOI{10.1145/XXXXXX.XXXXXX} 
\acmISBN{978-1-4503-XXXX-X/26/06}
\acmSubmissionID{1459}    

\renewcommand\footnotetextcopyrightpermission[1]{}

\ccsdesc[500]{Information systems~Multimedia information systems}
\keywords{Ambiguity Detection, 3D Scene Understanding, Embodied AI}

\title{3D Instruction Ambiguity Detection}

\author{Jiayu Ding, Haoran Tang, Hongbo Jin, Wei Gao, and Ge Li\footnotemark[1]}
\affiliation{
	\institution{School of Electronic and Computer Engineering, Peking University}
	\city{}       
	\country{}    
}
\begin{abstract}
	In safety-critical domains, linguistic ambiguity can have severe consequences; a vague command like \enquote{Pass me the vial} in a surgical setting could lead to catastrophic errors. Yet, most embodied AI research overlooks this, assuming instructions are clear and focusing on execution rather than confirmation. To address this critical safety gap, we are the first to define 3D Instruction Ambiguity Detection, a fundamental new task where a model must determine if a command has a single, unambiguous meaning within a given 3D scene. To support this research, we build Ambi3D, the large-scale benchmark for this task, featuring over 700 diverse 3D scenes and around 22k instructions. Our analysis reveals a surprising limitation: state-of-the-art 3D  Large Language Models (LLMs) struggle to reliably determine if an instruction is ambiguous. 
	To address this challenge, we propose AmbiVer, a two-stage framework that collects explicit visual evidence from multiple views and uses it to guide an vision-language model (VLM) in judging instruction ambiguity.
	Extensive experiments demonstrate the challenge of our task and the effectiveness of AmbiVer, paving the way for safer and more trustworthy embodied AI. Code and dataset available at https://jiayuding031020.github.io/ambi3d/.
\end{abstract}
\begin{document}
	\maketitle
	\footnotetext[1]{*Corresponding author.}
\section{Introduction}
\label{sec:intro}
\begin{figure}[t]
	\centering
	\includegraphics[width=1.0\columnwidth]{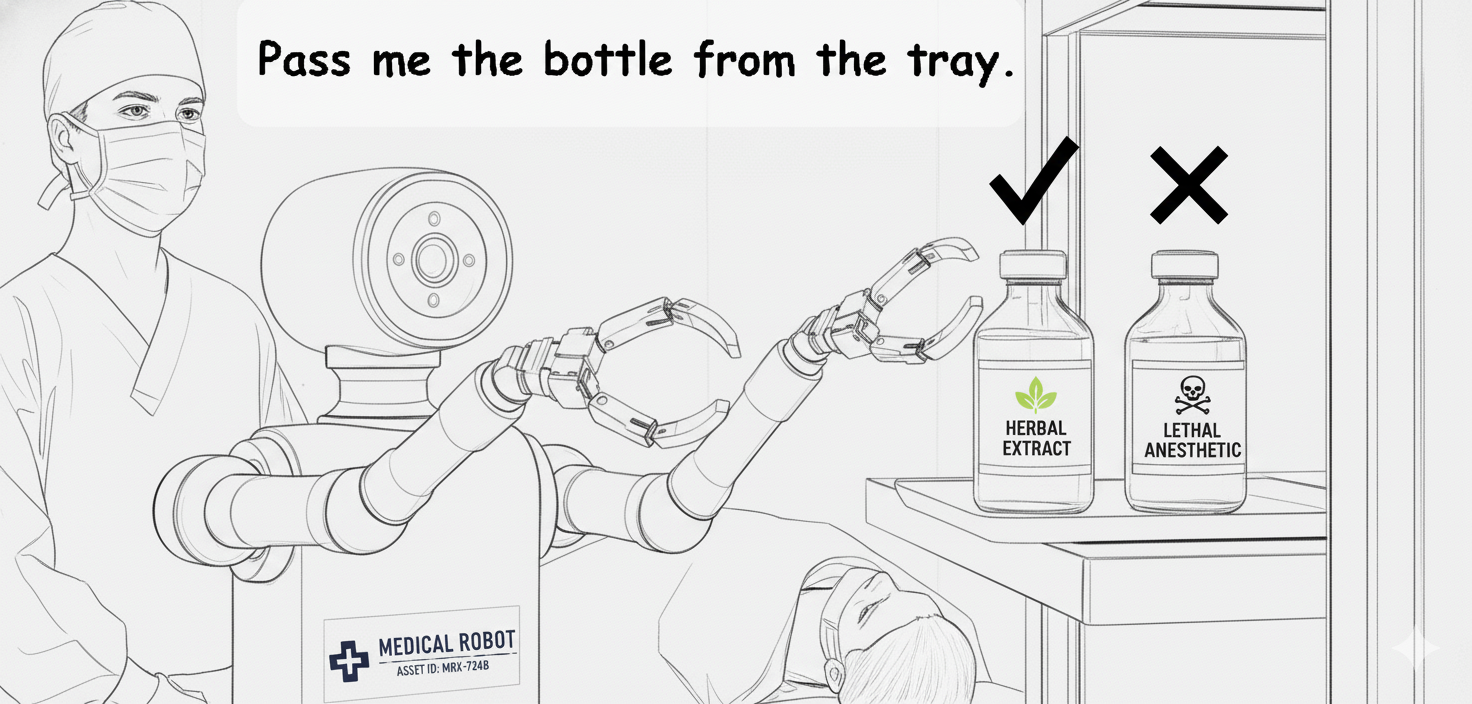}
	\caption{This high-stakes scenario highlights a critical safety challenge where an ambiguous instruction forces a robot to choose between a harmless substance and a lethal one.}
	\label{fig:ambiguity_example}
\end{figure}

The reliability of an agent interacting with the physical world heavily depends on its precise understanding of human instructions. Figure~\ref{fig:ambiguity_example} illustrates a safety-critical scenario: a surgeon instructs a robot to \enquote{Pass me the vial from the tray} when both a benign herbal extract and a lethal anesthetic are present. A system incapable of recognizing this instructional ambiguity might arbitrarily select an object, leading to potentially catastrophic consequences. This challenge extends far beyond the operating room. Similar safety risks arising from linguistic ambiguity are prevalent across numerous domains, including home services, industrial automation, and augmented reality operations. Therefore, a trustworthy intelligent system must actively identify and resolve such ambiguities before executing any physical action.

Despite the critical importance of this challenge, the research focus in embodied intelligence has predominantly centered on \enquote{grounding} language in vision and subsequent \enquote{execution}. While this paradigm has achieved remarkable success in many tasks, its inherent unambiguous instruction assumption introduces significant latent risks. Visual Question Answering (VQA), for instance, has significantly advanced multimodal understanding, but its models are often optimized on datasets where questions are presumed to have a single, verifiable answer. For example, in a smart home context, confronted with the question, \enquote{Was the stove in the kitchen turned off when I left?} a VQA model might observe that the main stovetop is off and answer \enquote{Yes}, overlooking a portable induction cooktop still operating in a corner. This affirmative answer creates a false sense of security, rooted in the model's inability to recognize that the scope of \enquote{the stove in the kitchen} is itself ambiguous. Fundamentally, these models are designed to find the \enquote{right answer} to an input assumed to be valid, lacking an intrinsic mechanism to identify when the input itself is a \enquote{bad question}. This challenge is exacerbated in complex 3D environments, where referents may be partially occluded and their spatial relationships are often view-dependent. A truly reliable system must possess the ability to recognize such ambiguity and proactively seek clarification, rather than blindly guessing.

This tendency toward blind guessing reflects a longstanding bias in embodied AI research: an emphasis on the correctness of instruction execution while overlooking the executability of the instruction itself.
Recent studies have begun to address linguistic ambiguities arising in human–agent interaction, proposing two broad classes of resolution strategies.
The first is \textit{passive resolution}, which makes a \enquote{best guess} under uncertainty: the agent either infers the most plausible target from contextual priors~\cite{LMCR_cite, MMC-GAN_cite} or executes a tentative action and relies on human feedback for correction~\cite{FindThis_cite, ZIPON_cite}.
The second is \textit{active clarification}~\cite{KnowNo_cite, CoIN_cite}, where the agent proactively queries the user when recognizing low confidence or uncertain actions.
However, both paradigms base ambiguity judgments on the model’s \textit{internal subjective state}. As a result, a model may express high confidence in an objectively ambiguous instruction yet hesitate over a clear one.
Consequently, the agent lacks the foundational ability to first ask: \textit{``Is this instruction objectively ambiguous given this specific 3D environment?''}

To systematically address this fundamental safety problem, we are the first to formally define the new task of \textbf{3D Instruction Ambiguity Detection}. The task requires a model to take a 3D scene and an natural language instruction as input, and to determine whether the instruction is ambiguous.
To facilitate systematic research on this new task, we construct \textbf{Ambi3D}, the large-scale benchmark featuring a meticulously human-annotated set of instructions that capture complex referential and execution ambiguities grounded in real-world scenes.
However, our analysis reveals a surprising limitation in existing methods: state-of-the-art 3D LLMs and Video LLMs struggle to reliably determine whether an instruction is ambiguous.
To overcome this limitation, we propose \textbf{Ambi}guity \textbf{Ver}ifier (AmbiVer), a two-stage framework that decouples scene perception from logical reasoning.  Its perception stage converts the raw scene and instruction into a set of structured evidence. The reasoning stage then passes this evidence to a zero-shot VLM for logical adjudication. Ultimately, AmbiVer establishes a new state-of-the-art on the Ambi3D benchmark, delivering superior detection capabilities with remarkable efficiency.

In summary, our contributions are as follows:
\begin{itemize}
	\item We define and formalize the critical safety task of 3D Instruction Ambiguity Detection.
	\item We build \textbf{Ambi3D}, a large-scale benchmark for this task, featuring $\sim$22k human-annotated instructions grounded in over 700 diverse real-world 3D scenes.
	\item We propose \textbf{AmbiVer}, a novel two-stage framework that decouples perception from logical reasoning, leveraging a VLM for zero-shot, evidence-based adjudication.
\end{itemize}

\section{Related Works}
\label{sec:formatting}
\textbf{Linguistic Ambiguity}
Linguistic ambiguity is a fundamental challenge in Natural Language Processing (NLP), stemming from words or phrases corresponding to multiple meanings~\cite{berry2004ambiguity, yadav2021comprehensive}.
Prior work has primarily focused on ambiguity arising from language-internal factors, branching into several key directions.
One direction is lexical ambiguity~\cite{petho2001polysemy, tabanakova2021term, abeysiriwardana2024survey}, which refers to a string of sounds or characters corresponding to multiple lexical or semantic interpretations.
Another research avenue is syntactic ambiguity~\cite{abeysiriwardana2024survey}.
This ambiguity arises not from individual words but from the grammatical structure of a sentence or phrase, leading to multiple valid parsing interpretations for the entire sentence.
Additionally, other types of ambiguity, such as semantic ambiguity~\cite{poesio1995semantic} and phonological ambiguity~\cite{frost1990phonological}, have also been explored.
However, this work is different from our research, as it concentrates on language-internal ambiguity rooted in lexicon or syntax.
In contrast, we focus on grounded instructional ambiguity, which is a property jointly determined by the instruction and the 3D scene.

\noindent \textbf{Ambiguity Resolution in Embodied AI.}
Although prior work in embodied AI has recognized the issue of linguistic ambiguity, research has largely focused on resolving rather than detecting it. Existing approaches generally fall into two paradigms.
The first, \textit{passive resolution}, includes both explicit and implicit forms. Explicit methods adopt a human-in-the-loop scheme~\cite{FindThis_cite, ZIPON_cite}, where the agent executes a ``best-guess'' action and relies on human feedback for correction. Implicit methods, by contrast, bypass explicit clarification by inferring missing information from context~\cite{LMCR_cite}, predicting the most plausible target~\cite{MMC-GAN_cite}, or exploiting auxiliary modalities such as gestures~\cite{M2Gestic_cite}.
The second, \textit{active clarification}~\cite{KnowNo_cite, CoIN_cite}, allows the agent to proactively query the user when facing low confidence or uncertainty in its action plan, thereby seeking explicit input to resolve ambiguity.
Furthermore, their associated benchmarks usually measure downstream task success (e.g., navigation) but not the correctness of the clarification decision itself.
In contrast, our work formalizes the upstream task of objective ambiguity detection, enabling an agent to adjudicate ambiguity based on explicit, factual 3D scene evidence.

\noindent \textbf{Open-Vocabulary 3D Scene Understanding}
The field of open-vocabulary 3D scene understanding seeks to interpret and interact with 3D environments using free-form natural language.
This field encompasses several key tasks, notably 3D Visual Question Answering (VQA), which requires generating answers to questions about a scene~\cite{azuma2022scanqa, zhu20233d, zhang2023multi3drefer}, and 3D Referring Expression Comprehension (REC), which focuses on locating objects from textual descriptions~\cite{qiao2020referring, kamath2021mdetr, ding2025polysemous, ReferSplat, liu2017referring}.
Across these tasks, the predominant trend has been a shift from task-specific expert models to versatile, general-purpose 3D LLMs~\cite{hong20233d, huang2024chat, zheng2025video, zhi2025lscenellm, zhu2025llava}.
These 3D LLMs align 3D features with the LLM embedding space, unlocking strong reasoning capabilities.
Both expert models and advanced 3D LLMs are architecturally biased to \enquote{find the best answer} or \enquote{locate the best match}, operating on the implicit assumption that the instruction is clear and unambiguous.
Our work addresses this gap by formalizing objective ambiguity detection as a crucial prerequisite for safe and reliable 3D scene interaction.

\section{Task Formulation}
In this section, we formalize our task by defining 3D instruction ambiguity from an execution-oriented perspective and characterizing its core types (\S\ref{sec:types_of_ambiguity}). Building upon this foundation, we present the formal task definition (\S\ref{sec:task_definition}).

\subsection{Formalizing 3D Instruction Ambiguity}
\label{sec:types_of_ambiguity}

Diverging from traditional NLP's focus on intrinsic textual ambiguity, our research investigates scene-grounded instructional ambiguity as it pertains to an agent's task execution in the 3D physical world. This execution-centric focus necessitates that we define ambiguity from a pragmatic perspective:
\enquote{\textit{An instruction is ambiguous if insufficient information or vague descriptions compel an agent to rely on hazardous guesswork or request clarification to ensure safe completion}}. This allows us to focus on critical failures in human-robot interaction while filtering out acceptable vagueness. We categorize these critical issues into two primary classes: Referential Ambiguity and Execution Ambiguity.

Referential Ambiguity arises when an instruction fails to allow an agent to isolate a single, well-defined set of target objects.
We identify three primary types of this ambiguity: Instance Ambiguity, Attribute Ambiguity, and Spatial Ambiguity.
\textbf{1)} Instance Ambiguity occurs when an instruction uses a general class name (e.g., \enquote{the cup} or \enquote{the chair}) without any distinguishing features, but multiple objects of that class exist in the scene. 
\textbf{2)} Attribute Ambiguity arises  
from the use of subjective (e.g., \enquote{the nice book}) or relative adjectives (e.g., \enquote{the large chair}) that are not explicitly unique (e.g., "largest"). The vagueness of these terms can result in multiple objects satisfying the description. 
\textbf{3)} Spatial Ambiguity stems from observer-dependent spatial terms (e.g., \enquote{to the left of}), where the instruction's correct interpretation changes based on the agent's or user's viewpoint.

In contrast to referential issues, Execution Ambiguity occurs when the target object is clear, but the action verb itself has multiple plausible and mutually exclusive interpretations (e.g., \enquote{deal with the cup} could mean to set it upright, move it, or discard it).

Based on this, we define an instruction as unambiguous only if it maps precisely and uniquely to a single target object or a fixed set of target objects, and its core action entails no conflicting interpretations.

\subsection{Task Definition}
\label{sec:task_definition}
We introduce the task of 3D Instruction Ambiguity Detection. We formally define this as a binary classification problem, where a model $\mathcal{F}$ learns the mapping $\mathcal{F}: (S, T) \mapsto y$. The inputs consist of a 3D scene representation $S$ and an natural language instruction $T$. The model must output a single binary label $y \in \{\text{Unambiguous, Ambiguous}\}$. The core challenge is to compel $\mathcal{F}$ to replace the conventional forced-choice selection paradigm with a rigorous, quantity-aware perceptual and reasoning process to verify instructional uniqueness.

\section{Benchmark}
To facilitate the systematic study of 3D Instruction Ambiguity Detection, we construct \textbf{Ambi3D}, the first benchmark for this task, built upon the ScanNet dataset.
In this section, we detail the dataset's acquisition pipeline, annotation process, and statistics (\S\ref{sec:dataset}) and specify the evaluation protocol (\S\ref{sec:metrics}). 

\subsection{Dataset}
\label{sec:dataset}
\noindent\textbf{Instruction Acquisition Pipeline}
We designed a meticulous instruction acquisition pipeline to ensure the dataset's authenticity, diversity, and challenge. The pipeline consists of three main components: 
\textbf{1)}~Grounded Instructions: We leverage the high-quality, human-annotated question-answer (QA) pairs from the ScanQA dataset. We employ an LLM-based framework to automatically convert these questions (e.g., \enquote{What is the tallest object on the table?}) into semantically equivalent, executable instructions (e.g., \enquote{Please pick up the tallest object on the table}). This provides a robust, real-world semantic foundation for a subset of our data. 
\textbf{2)}~Synthetic Ambiguous Instructions: To systematically cover the ambiguity types defined in Section~\ref{sec:types_of_ambiguity}, we designed a prompt-engineered LLM generation process based on detailed object metadata from ScanNet scenes. This process uses specific templates to target the generation of four ambiguity types: Instance, Attribute, Spatial, and Action Ambiguity. 
\textbf{3)}~Hard Negative Instructions: To evaluate model robustness and prevent superficial heuristics, we also constructed a set of special unambiguous instructions. These instructions appear ambiguous on the surface (e.g., referring to a \enquote{chair} in a scene with multiple chairs) but are implicitly disambiguated by a unique qualifier (e.g., \enquote{the chair by the window}). To ensure the quality and subtlety of these samples, this subset was entirely human-authored by experts.

\noindent\textbf{Annotation and Quality Control}
To ensure high-fidelity labels, all generated instructions underwent a multi-stage verification process conducted by 12 trained annotators, all possessing backgrounds in 3D domains. To ensure proficiency, annotators were provided with comprehensive guidelines detailing ambiguity definitions and boundary cases, and were required to pass a qualification test before beginning. Our annotation process consisted of the following three stages:
\textbf{1)}~Data Cleaning and Filtering: We first applied an automated script to identify and remove instructions that were exact duplicates within the same scene. Subsequently, human annotators performed a manual review to filter out any remaining instructions that were grammatically incorrect, semantically nonsensical, or irrelevant to the 3D scene context.
\textbf{2)}~Core Ambiguity Annotation: Following cleaning, each valid instruction entered the core annotation stage, where it was independently assigned to three different annotators. Annotators provided two labels based on the 3D scene context: a primary binary label (Unambiguous or Ambiguous) and, if ambiguous, a secondary sub-type classification (e.g., Instance, Attribute, Action).
\textbf{3)}~Consistency Check and Final Selection: Finally, we employed a strict unanimous agreement protocol for final selection. To ensure maximum reliability for the primary task, we retained only those instructions where all three annotators reached a unanimous agreement on the binary label. 
For the retained ambiguous samples, the final sub-type label was determined by a majority vote. 
If at least two of the three annotators agreed on a specific sub-type, that type was assigned. 
However, in cases where no majority was reached (i.e., all three annotators selected a different sub-type), the sample was discarded from the final dataset.
This protocol eliminates arbitrary bias by ensuring every assigned sub-type is backed by a consensus of at least two annotators.

\noindent\textbf{Dataset Statistics and Splits}
The Ambi3D benchmark contains 22,081 instructions grounded in 703 unique indoor scenes from ScanNet. It features 10,480 Unambiguous instructions (47.5\%) and 11,601 Ambiguous instructions (52.5\%). The ambiguous instructions are categorized by type: 5,333 Instance (46.0\%), 2,302 Action (19.8\%), 2,216 Attribute (19.1\%), and 1,750 Spatial (15.1\%). Further dataset details and examples are available in the appendix.

These instructions originate from our three acquisition pipelines: 8,224 Grounded Instructions (37.2\%), 7,522 Synthetic Ambiguous Instructions (34.1\%), and 6,335 Hard Negative candidates (28.7\%). All instructions, regardless of their pipeline source, received their final ground-truth label from the identical human annotation process.

For experiments, we split the dataset at the scene level into 649 scenes for training and 54 for testing. The training set comprises 19,950 instructions (90.3\%), with 10,528 ambiguous (52.8\%) and 9,422 unambiguous (47.2\%) samples. The test set contains the remaining 2,131 instructions (9.7\%), with 1,073 ambiguous (50.4\%) and 1,058 unambiguous (49.6\%), ensuring an unbiased evaluation.

Linguistically, instructions have an average length of 8.08 words. Nearly half (49.35\%) fall into a medium-complexity range (6--10 words), with the remainder balanced between simple ($ \le $5 words, 28.11\%) and complex ($ > $10 words, 22.53\%), testing model robustness across diverse syntactic structures.
\subsection{Evaluation Metrics}
\label{sec:metrics}
We evaluate this task as a binary classification problem. For a comprehensive assessment of overall performance, we report standard Accuracy (Acc.), Precision (Prec.), Recall (Rec.) and the macro-averaged F1-Score (F1). Furthermore, to provide a fine-grained diagnostic, we report the Accuracy breakdown for each specific category: Instance, Attribute, Spatial, Action, and Unambiguous.

\section{Method}
We propose \textbf{AmbiVer}, the first unified framework for 3D instruction ambiguity detection.
In this section, we first outline the overall system architecture (\S\ref{sec:overall_arch}). 
We then detail its two core components: the perception engine (\S\ref{sec:perception_engine}) for extracting visual evidence, and the reasoning engine (\S\ref{sec:reasoning_engine}) for adjudicating ambiguity.

\begin{figure*}[t]
	\centering
	\includegraphics[width=\linewidth]{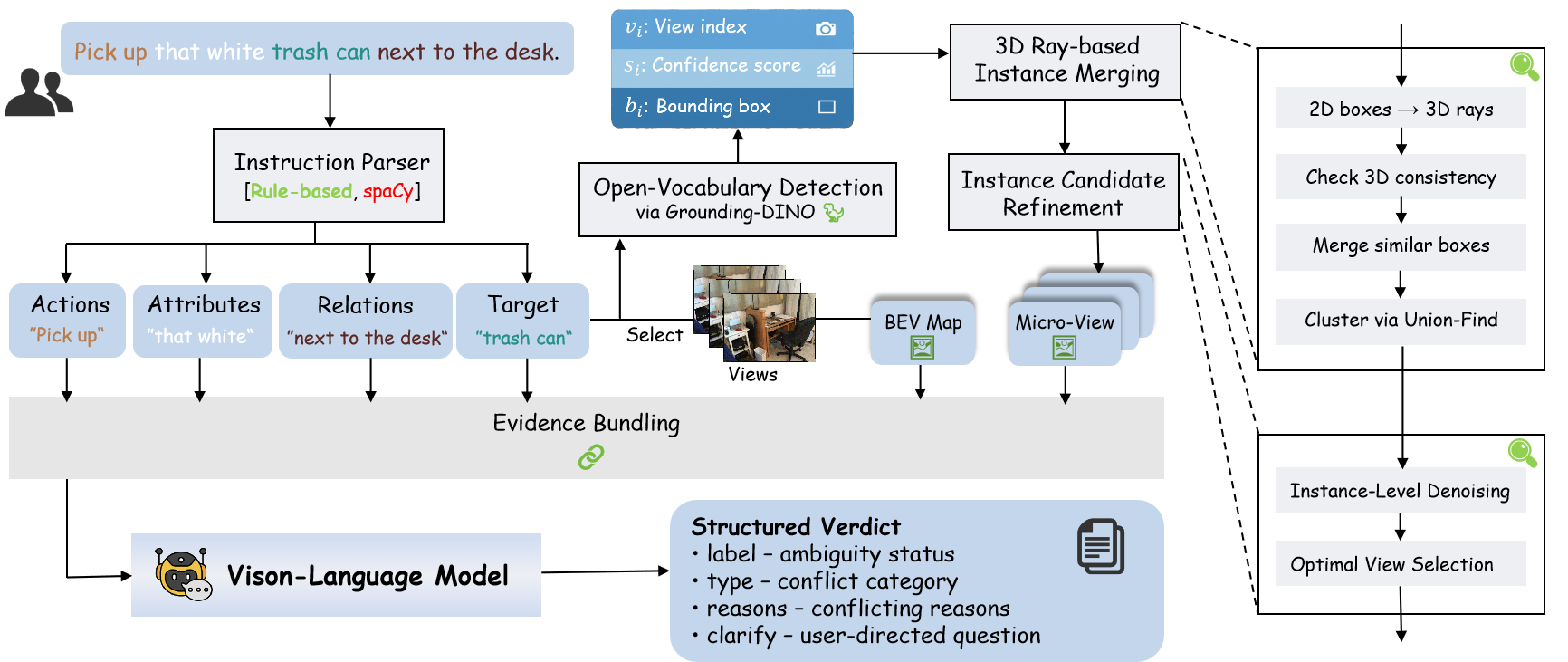}
	\caption{Overview of the \textbf{AmbiVer} framework.
		AmbiVer is a two-stage system composed of a perception engine and a reasoning engine. The perception stage parses an instruction into action, attribute, relation, and target components, employs an open-vocabulary grounding method to detect 2D candidates across views, and integrates them into 3D instances via ray-based fusion followed by refinement. It also generates a BEV  map for scene-level context. The reasoning stage then performs multimodal evidence bundling and leverages a VLM to assess instruction ambiguity, outputting a structured verdict.}
	\label{fig:ambiver_overview}
\end{figure*}

\subsection{Overall Architecture}
\label{sec:overall_arch}

The AmbiVer pipeline, illustrated in Figure~\ref{fig:ambiver_overview}, is a decoupled, two-stage framework: a perception engine (\S\ref{sec:perception_engine}) and a reasoning engine (\S\ref{sec:reasoning_engine}).
The pipeline operates on an egocentric video stream $\mathcal{V} = \{\mathcal{I}_t\}_{t=1}^{N}$, corresponding camera extrinsics $\mathcal{E} = \{\mathcal{E}_t\}_{t=1}^{N}$, and a natural language instruction $T$.
The Perception Engine is responsible for processing these raw inputs, converting the video stream and instruction into a set of structured, multimodal evidence. 
This evidence is then passed to the reasoning engine, which employs a zero-shot VLM to perform logical deliberation on the evidence, ultimately outputting a structured $\texttt{verdict}$ that adjudicates the instruction's ambiguity.

\subsection{Perception Engine: From Pixels to Evidence}
\label{sec:perception_engine}
The perception engine must process two challenging and unstructured inputs: \textbf{1)} the free-form language instruction $T$, which must be parsed, and \textbf{2)} the egocentric video stream $\mathcal{V}$, which offers only partial and localized observability.
To convert this raw data into actionable evidence, we design a multi-track pipeline.
This section details the three core sub-modules: 
\textbf{1)} Instruction Decoupling, which parses $T$ into a structured representation of its core components;
\textbf{2)} Global Feature Acquisition, which aggregates $\mathcal{V}$ into a unified, allocentric Bird’s-Eye View (BEV) map $\mathcal{I}_{\text{bev}}$ for scene-level context; and \textbf{3)} Detailed Feature Acquisition, which localizes object candidates from $\mathcal{V}$ and resolves the multi-view redundancies to identify all potential 3D instances.

\noindent\textbf{Instruction Decoupling}
We employ \textit{spaCy} to perform lexical analysis (tokenization and part-of-speech tagging) and syntactic analysis (dependency parsing and named entity recognition) on the input instruction $T$. 
This process converts the free-form text $T$ into a structured key–value representation containing essential elements such as the action, target (denoted as $Q_{\text{t}}$), attributes, and relations. 
Once parsed, the instruction guides the perception engine along two complementary tracks: constructing a global allocentric map and identifying fine-grained local instances.

\noindent\textbf{Global Feature Acquisition}
The perception engine synthesizes raw visual inputs into actionable evidence. 
However, the egocentric video stream \( \mathcal{V} = \{\mathcal{I}_t\}_{t=1}^{N} \) provides only a limited field of view, 
capturing partial and localized observations. 
To overcome this constraint and achieve a global understanding of the environment, 
the engine transforms the first-person views into an allocentric representation of the scene. This process aggregates multi-view observations into a unified 3D point cloud \( \mathcal{P} \), 
reconstructed from the video frames \( \mathcal{V} \) and their corresponding camera poses 
\( \mathcal{E} = \{\mathcal{E}_t\}_{t=1}^{N} \) through a reconstruction pipeline \( \mathcal{R} \):
\begin{equation}
	\mathcal{P} = \mathcal{R}\!\left( \{ (\mathcal{I}_t, \mathcal{E}_t) \}_{t=1}^{N} \right).
\end{equation}
The point cloud \( \mathcal{P} \) encodes the full 3D geometry of the scene. 
To represent this geometry in a structured 2D form, we project \( \mathcal{P} \) into a BEV image 
\( \mathcal{I}_{\text{bev}} \) using a transformation \( \mathcal{T} \) 
defined by a fixed top-down camera extrinsic \( \mathcal{E}_{\text{top}} \in SE(3) \):
\begin{equation}
	\mathcal{I}_{\text{bev}} = \mathcal{T}(\mathcal{P}, \mathcal{E}_{\text{top}}).
\end{equation}
The BEV image \( \mathcal{I}_{\text{bev}} \) provides a compact and globally consistent spatial layout of the environment, 
serving as the foundation for subsequent spatial reasoning.

\noindent\textbf{Detailed Feature Acquisition}
\label{sec:detailed_features}
The core task of the perception engine is to identify and enumerate object instances that match the query target $Q_{\text{t}}$ across multiple views, 
and to determine the most representative view for each instance. 
To this end, we design a multi-stage process that progressively refines the visual evidence.

First, processing the entire video stream $\mathcal{V}$ is computationally prohibitive. 
We therefore apply an adaptive keyframe selection strategy to downsample the stream to a target frame count $N_{\text{t}}$. 
The algorithm iteratively scans the sequence, retaining a new keyframe only when its pose deviation from the last selected frame 
exceeds the translational ($\tau_t$) or rotational ($\tau_r$) thresholds. 
To converge the keyframe count $N_c$ to the target $N_{\text{t}}$, the thresholds $\tau_t$ and $\tau_r$ are iteratively adjusted. In each iteration, both thresholds are multiplicatively scaled by a factor $\alpha_i > 1$ if $N_c > N_{\text{t}}$ or by $\alpha_d < 1$ if $N_c < N_{\text{t}}$, until $N_c$ falls within a predefined tolerance window around $N_{\text{t}}$.
This procedure produces a compact yet diverse set of keyframes $\{I_v\}_{v=1}^{N_{\text{t}}}$ and their corresponding poses $\{\mathcal{E}_v\}_{v=1}^{N_{\text{t}}}$.

Given these keyframes, the next step is to localize potential object candidates. 
For each $I_v$, we employ the pre-trained open-vocabulary detector Grounding DINO~\cite{liu2024grounding} 
with the query text $Q_{\text{t}}$. This process yields an aggregated set of 2D detections 
$\mathcal{D}=\{d_i=(v_i, b_i, s_i)\}_{i=1}^{M}$, 
where $v_i$ is the index of the keyframe $I_{v_i}$ where the detection was found, $b_i$ is its 2D bounding box, and $s_i$ is its detection confidence score. 
Because the same object can appear in multiple views, $\mathcal{D}$ consequently contains redundant detections.

To unify redundant multi-view detections $\mathcal{D}$ into consistent 3D instances, we construct a connectivity graph where each detection $d_i$ (back-projected as ray $\text{Ray}_i=(o_i, r_i)$) serves as a node. An edge is formed between two nodes $d_i$ and $d_j$ (from distinct views $v_i \neq v_j$) if they satisfy three geometric constraints: \textbf{1)} their minimum ray distance is less than $\epsilon_d$; \textbf{2)} their ray angle is within $[\theta_{a, \text{min}}, \theta_{a, \text{max}}]$; and \textbf{3)} their bounding box area ratio, $\frac{\min(\text{area}(b_i), \text{area}(b_j))}{\max(\text{area}(b_i), \text{area}(b_j))}$, exceeds $\sigma_s$. An efficient Union-Find algorithm is applied to extract the connected components $\{\mathcal{G}_k\}_{k=1}^{N_g}$, each representing a unified 3D instance.

Subsequently, we assign a group-level reliability score $S_k$ to each group $\mathcal{G}_k$ 
based on the area-weighted average confidence of its constituent detections $d_i \in \mathcal{G}_k$:
\begin{equation}
	S_k = 
	\frac{\sum_{d_i \in \mathcal{G}_k} s_i \cdot \text{area}(b_i)}
	{\sum_{d_i \in \mathcal{G}_k} \text{area}(b_i)}.
	\label{eq:group_score}
\end{equation}
Instances are ranked by $S_k$, and only the top $K$ are retained to eliminate 
low-confidence or fragmented hypotheses. 
For each selected instance $\mathcal{G}_k$, a single representative detection 
$d_k^*=(v_k^*,b_k^*,s_k^*)$ is chosen from its members $d_i \in \mathcal{G}_k$ by maximizing a composite score $f(d_i)$.
This score is defined as the product of three components:
\begin{equation}
	f(d_i) = s_i \cdot w_{\text{vis}}(d_i) \cdot w_{\text{bnd}}(d_i)
	\label{eq:rep_score}
\end{equation}
where $s_i$ is the raw detection confidence, $w_{\text{vis}}(d_i) = \text{area}(b_i) / \text{area}(I_{v_i})$ measures visibility within the keyframe $I_{v_i}$, and $w_{\text{bnd}}(d_i)$ is a boundary penalty set to $\gamma$ if $b_i$ is within $\delta$ pixels of any image border, and $1.0$ otherwise. 
The detection $d_i \in \mathcal{G}_k$ that maximizes this score is selected as the representative $d_k^*$.

Finally, this process outputs the set of $K$ instance candidates $\mathcal{C}$, defined as:
\begin{equation}
	\mathcal{C} = \{ (I_{v_k^*}, b_k^*, S_k, |\mathcal{G}_k|) \}_{k=1}^K
	\label{eq:candidate_set}
\end{equation}
where each candidate $\mathcal{C}_k$ consists of its representative image $I_{v_k^*}$ (from the keyframe set $\{I_v\}$), bounding box $b_k^*$, group reliability score $S_k$, and the group's cardinality $|\mathcal{G}_k|$ (i.e., its cross-view detection count).
This set $\mathcal{C}$, combined with the BEV map $\mathcal{I}_{\text{bev}}$, forms the complete visual evidence for the reasoning engine.

\subsection{Reasoning Engine: From Evidence to Verdict}
\label{sec:reasoning_engine}
The reasoning engine evaluates the structured evidence from the perception pipeline against the instruction's semantic constraints. This process aggregates the evidence into a unified $\texttt{Dossier}$ and uses a zero-shot VLM to adjudicate it, producing the final, interpretable $\texttt{verdict}$.

\noindent\textbf{Structured Evidence Bundling}
\label{sec:bundling}
Before reasoning, we aggregate the outputs from the perceptual pipeline into a structured evidence package, referred to as the $\texttt{Dossier}$. 
This package integrates linguistic, geometric, and visual information into a coherent representation for the VLM. It organizes the multimodal data into three complementary components: 
\textbf{1)} Linguistic Context, which contains the raw instruction $T$ and its parsed components, including the query target $Q_{\text{t}}$, attributes, and relational constraints; 
\textbf{2)} Global Spatial Context, represented by the top-down BEV map $\mathcal{I}_{\text{bev}}$, providing a holistic view of the environment; and 
\textbf{3)} Local Instance Evidence, corresponding to the set of $K$ unified instance candidates $\mathcal{C}$. 
Each instance $\mathcal{C}_k$ is associated with its representative image $I_{v_k^*}$, bounding box $b_k^*$, reliability score $S_k$, and the group's cardinality $|\mathcal{G}_k|$.

\noindent\textbf{VLM as a Zero-Shot Adjudicator}
\label{sec:vlm_adjudicator}
We employ a general-purpose VLM as a zero-shot logical adjudicator, tasked with evaluating whether the perceived scene satisfies the semantic constraints imposed by the instruction. 
To facilitate such reasoning, \texttt{Dossier} (\S\ref{sec:bundling}) is encapsulated within a carefully designed multimodal prompt. This prompt guides the VLM's evidence-based verification by combining the evidence with an explicit execution-oriented ambiguity criterion, which is detailed in the Appendix.
The model performs zero-shot reasoning over the full prompt to generate a structured $\texttt{verdict}$, 
consisting of four fields: a binary ambiguity $\texttt{label}$ (e.g., ``Ambiguous'' or ``Unambiguous''), 
a set of detected ambiguity types (e.g., ``Instance'', ``Spatial''), 
a concise textual explanation, 
and an optional clarification query for disambiguation. 
This structured formulation ensures that the results are directly parsable for downstream evaluation while maintaining interpretability.

\section{Experiments}
In this section, we first describe implementation details (\S\ref{sec:implementation_details}). We then evaluate our framework via quantitative (\S\ref{sec:quant_analysis}) and qualitative (\S\ref{sec:qualitative}) analyses. We further assess its cross-dataset generalization (\S\ref{sec:generalization}) and conduct ablation studies (\S\ref{sec:abs}) to validate our design.

\begin{table*}[t]
	\centering
	\caption{
		Main performance comparison on the Ambi3D benchmark in the zero-shot setting. Evaluation is performed on the \textit{entire} dataset for a comprehensive overview. For Video LLMs, all results are reported as 0-shot accuracy on 7B or 8B size models. For AmbiVer, the reported $\sim$4.56 frames represent the average number of distilled visual inputs (representative instance views and the BEV map) ultimately fed into the VLM. 
	}
	
	\label{tab:main_results}
	\renewcommand{\arraystretch}{0.96}
	\setlength{\aboverulesep}{2pt}
	\setlength{\belowrulesep}{2pt}
	\setlength{\tabcolsep}{8pt}
	\begin{tabular}{l c c cc ccccc}
		\toprule
		\multirow{2}{*}{\textbf{Method}} & \multirow{2}{*}{\textbf{Venue}} & \multirow{2}{*}{\textbf{Frames}} & \multicolumn{2}{c}{\textbf{Overall Metrics}} & \multicolumn{5}{c}{\textbf{Acc. Breakdown by Type}} \\
		\cmidrule(lr){4-5} \cmidrule(lr){6-10} 
		& & & Acc. $\uparrow$ & Macro-F1 $\uparrow$ & Instance & Attribute & Spatial & Action & Unamb. \\
		\midrule
		
		\multicolumn{10}{l}{\textbf{\textit{3D LLMs}}} \\
		3D-LLM~\cite{hong20233d} & NIPS'23 & - & 49.16 & 41.97 & 13.59 & 14.85 & 11.71 & 12.25 & 88.87 \\
		Chat-Scene~\cite{huang2024chat} & NIPS'24 & - & 47.73 & 42.84 & \underline{18.94} & 11.82 & \underline{17.14} & \underline{20.29} & 81.12 \\
		Video-3D LLM~\cite{zheng2025video} & CVPR'25 & - & \underline{50.10} & \underline{43.11} & 12.94 & \underline{35.79} & 7.03 & 2.35 & 89.72 \\
		LSceneLLM~\cite{zhi2025lscenellm} & CVPR'25 & - & 48.78 & 39.46 & 9.36 & 11.01 & 6.06 & 8.86 & \underline{92.74} \\
		Robin3D ~\cite{kang2025robin3d} & ICCV'25 & - & 48.50 & 38.26 & 6.98 & 9.66 & 5.26 & 7.78 & \textbf{94.01} \\
		LLaVA-3D~\cite{zhu2025llava} & ICCV'25 & - & \textbf{64.21} & \textbf{63.63} & \textbf{59.55} & \textbf{95.62} & \textbf{62.86} & \textbf{90.96} & 54.29 \\
		
		\midrule
		\multicolumn{10}{l}{\textbf{\textit{Video LLMs}}} \\
		LLaVA-Video~\cite{zhang2024llava} & ArXiv'24 & 8 & \underline{57.88} & 50.60 & \textbf{90.36} & \textbf{97.16} & \textbf{88.06} & \textbf{91.96} & 20.52 \\
		LLaVA-NeXT-Video & - & 8 & 48.14 & 34.81 & 3.13 & 5.87 & 0.29 & 0.87 & \textbf{98.36} \\
		InternVL-3.5~\cite{wang2025internvl3} & ArXiv'25 & 8 & 55.89 & 50.24 & 21.00 & 39.26 & 10.86 & 11.60 & \underline{94.42} \\
		Qwen3-VL~\cite{bai2025qwen3} & ArXiv'25 & 8 & 56.76 & \underline{55.40} & 27.51 & 66.74 & 48.74 & 23.28 & 78.23 \\
		\rowcolor{gray!10} \textbf{AmbiVer} & -- & $\sim$4.56 & \textbf{66.16} & \textbf{66.15} & \underline{53.57} & \underline{87.55} & \underline{78.06} & \underline{58.60} & 67.72 \\
		\bottomrule
	\end{tabular}
\end{table*}

\subsection{Implementation Details.}
\label{sec:implementation_details}
We set the adaptive keyframe selection target to $N_{\text{target}}=100$. For instance unification, we use a ray distance threshold of $0.3$m, an angular limit of $60^{\circ}$, and a scale similarity of $0.2$. We retain the top $K=6$ instances for reasoning. The representative score (Eq.~\ref{eq:rep_score}) uses a boundary penalty $\gamma=0.5$ within $\delta=4$ pixels. The zero-shot reasoning is performed using Qwen-3-VL-8b-Instruct~\cite{yang2025qwen3} with a temperature of 0. 
For the Low-Rank Adaptation (LoRA) of 3D LLM baselines, we train for 3 epochs using the AdamW optimizer with a learning rate of $2\times10^{-4}$, a cosine learning rate schedule, a weight decay of $0.01$, and a 3\% warmup ratio. The LoRA rank is $r=8$, $\alpha=16$, and dropout is $0.1$.
We partition the official training set into a 90\% subset for training and a 10\% subset for validation.
During training, we select the model checkpoint that achieves the highest Acc. score on the validation set.
The test set was strictly held out and used only for the final, one-time evaluation.

\subsection{Quantitative Analysis}
\label{sec:quant_analysis}
We systematically evaluate existing 3D Large Language Models (3D LLMs), Video Language Models (Video LLMs), and our proposed AmbiVer framework on the Ambi3D benchmark to investigate their zero-shot understanding capabilities. The quantitative results are presented in Table~\ref{tab:main_results}. As existing baselines are not natively designed for this binary classification task, we employ specific prompts to constrain their outputs to either 0 (unambiguous) or 1 (ambiguous). To ensure a fair comparison, the prompts provided to the baselines are identical to those used for AmbiVer, aside from the output format constraints (see Appendix for details). However, even under strict guidance, some baselines occasionally generate free-form text (e.g., "The instruction is clear") rather than strictly adhering to the binary format requirement. To accurately extract the final predictions and maintain evaluation objectivity, we apply a robust rule-based parsing algorithm, detailed in the Appendix.

As shown in Table \ref{tab:mipnerf_results}, existing zero-shot baselines struggle with ambiguous scenarios. Lacking an explicit mechanism to verify instruction validity, end-to-end 3D LLMs fail to handle ambiguities reliably and exhibit severe biases. Specifically, when confronted with ungroundable commands, some models force unambiguous predictions and hallucinate, while others conservatively assume the instructions are invalid, leading to incorrect rejections. Moreover, while Video LLMs process temporal information, they lack precise 3D spatial perception, which hinders their ability to capture complex spatial occlusions or fine-grained multi-instance conflicts. In contrast, our AmbiVer framework significantly outperforms all baselines across all metrics. Furthermore, due to the adaptive keyframe selection in our perception module, AmbiVer achieves these superior results using substantially fewer visual frames than video models. This demonstrates that extracting structured, high-quality 3D evidence is more effective and efficient for disambiguation than feeding lengthy raw visual sequences into large multimodal models.
\begin{figure*}[t]
	\centering
	\includegraphics[width=\linewidth]{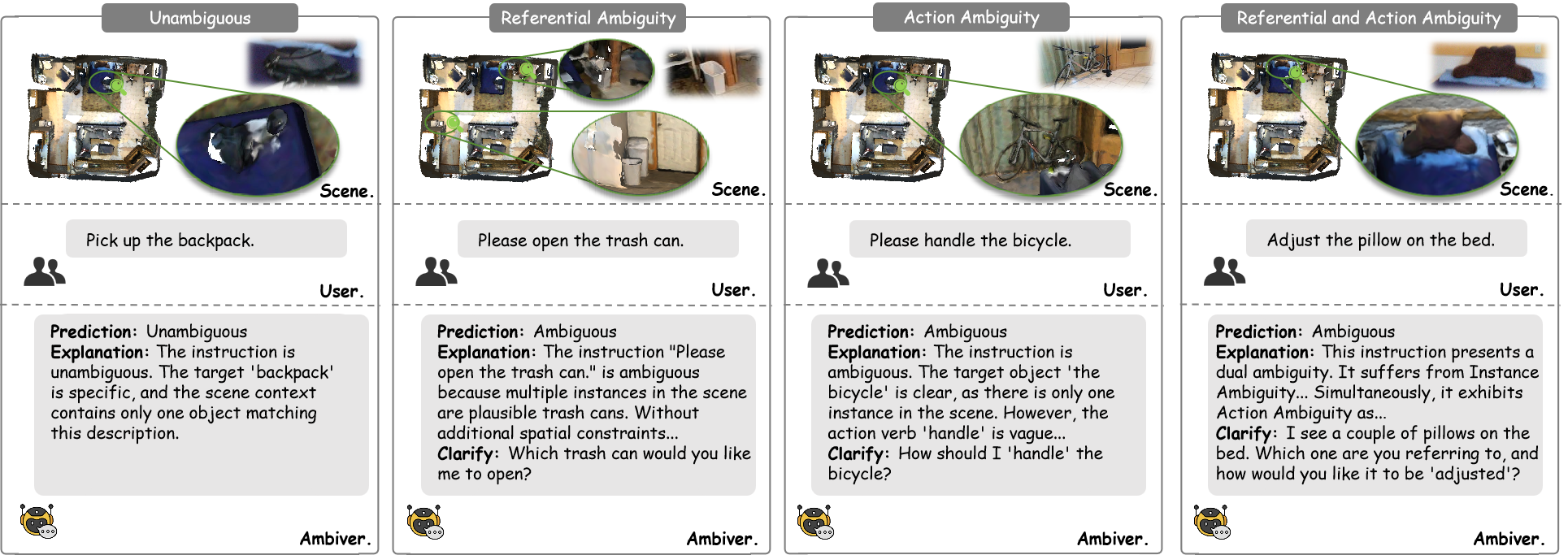}
	\caption{Qualitative results of our AmbiVer framework on the Ambi3D benchmark.}
	\label{fig:qualitative_examples}
\end{figure*}

To investigate whether in-domain data can alleviate the performance bottlenecks of baseline models in the zero-shot setting, we further conduct parameter-efficient fine-tuning (LoRA)~\cite{hu2022lora} on representative baselines using the Ambi3D training set, with results presented in Table~\ref{tab:lora_results}. Experiments show that supervised fine-tuning endows the baselines with a certain degree of pattern-fitting capability, leading to significant improvements across all evaluation metrics after learning from extensive in-domain ambiguous samples. However, even with sufficient fine-tuning, the performance of these baselines still falls short of AmbiVer. We attribute this performance bottleneck to the inherent limitations of existing end-to-end models (including 3D LLMs and Video LLMs) during the visual encoding stage. As long visual sequences are continuously compressed and abstracted within the network, fine-grained visual details closely related to user instructions suffer from severe, irreversible loss. Consequently, these models struggle to extract effective key clues from global visual features when handling ambiguity judgments that require precise local perception. In contrast, AmbiVer explicitly extracts fine-grained visual features highly relevant to the instruction via the perception module and transforms them into structured local evidence, thereby preventing the loss of critical information. Ultimately, AmbiVer requires an average of only 4 visual frames as input to achieve optimal performance. This proves that precise feature selection and decoupled architectures are more efficient and reliable than end-to-end feature compression for complex 3D scene cognition.

\begin{table}[t]
	\centering
	\caption{Comparison against LoRA fine-tuned baselines on the Ambi3D test set.}
	\label{tab:lora_results}
	\renewcommand{\arraystretch}{0.96}
	\setlength{\aboverulesep}{2pt}
	\setlength{\belowrulesep}{2pt}
	\setlength{\tabcolsep}{8pt}
	\begin{tabularx}{\linewidth}{l *{2}{C}}
		\toprule
		\textbf{Model} & \textbf{Acc.} $\uparrow$ & \textbf{Macro-F1} $\uparrow$ \\
		\midrule
		\multicolumn{3}{l}{\textbf{\textit{3D LLMs}}} \\
		3D-LLM~\cite{hong20233d} & 78.67 & 78.54 \\
		Chat-Scene~\cite{huang2024chat} & 78.74 & 79.17 \\
		Video-3D LLM~\cite{zheng2025video} & 79.63 &  \textbf{81.35} \\
		LSceneLLM~\cite{zhi2025lscenellm} & 77.24 & 77.05 \\
		Robin3D~\cite{kang2025robin3d} & \underline{80.76} & 78.73 \\
		LLaVA-3D~\cite{zhu2025llava} &  \textbf{80.95} & \underline{80.61} \\
		
		\midrule
		\multicolumn{3}{l}{\textbf{\textit{Video LLMs}}} \\
		LLaVA-Video~\cite{zhang2024llava} & 81.70 & 81.62 \\
		LLaVA-NeXT-Video & \underline{82.78} & \underline{82.68} \\
		InternVL-3.5~\cite{wang2025internvl3} & 76.91 & 76.86 \\
		Qwen3-VL~\cite{bai2025qwen3} & 82.17 & 80.21 \\
		\rowcolor{gray!10} \textbf{AmbiVer} & \textbf{84.80} & \textbf{84.54} \\
		\bottomrule
	\end{tabularx}
\end{table}

\subsection{Qualitative Analysis}
\label{sec:qualitative}
We present qualitative examples in Figure~\ref{fig:qualitative_examples} to illustrate our framework's efficacy. The figure highlights four representative cases: \textbf{1)} an unambiguous instruction (``Pick up the backpack'') where the model correctly identifies the unique referent; \textbf{2)} Referential Ambiguity (``...the trash can'') where the model detects multiple candidates and requests clarification; \textbf{3)} Action Ambiguity (``...handle the bicycle'') where the target is clear but the verb is flagged as vague; and \textbf{4)} Mixed Ambiguity (``...adjust the pillow...'') where our framework identifies both instance and action ambiguities, generating a comprehensive clarification question. These results confirm the efficacy of our decoupled approach.
\begin{table}[t]
	\centering
	\caption{
		Cross-dataset generalization accuracy on the Mip-NeRF 360 dataset in the zero-shot setting.
	}
	\label{tab:mipnerf_results}
	\renewcommand{\arraystretch}{0.96}
	\setlength{\aboverulesep}{2pt}
	\setlength{\belowrulesep}{2pt}
	\setlength{\tabcolsep}{8pt}
	\begin{tabular}{l ccc}
		\toprule
		\textbf{Method} & \textbf{Outdoor} & \textbf{Indoor} & \textbf{Average} \\
		\midrule
		
		\multicolumn{4}{l}{\textbf{\textit{3D LLMs}}} \\
		3D-LLM~\cite{hong20233d} & 44.90 & \underline{57.99} & 52.62 \\
		Chat-Scene~\cite{huang2024chat} & 47.13 & 56.12 & 52.43 \\
		Video-3D LLM~\cite{zheng2025video} & \textbf{72.22} & 57.18 & \underline{63.35} \\
		LSceneLLM~\cite{zhi2025lscenellm} & \underline{68.93} & \textbf{60.77} & \textbf{64.12} \\
		Robin3D~\cite{wang2025internvl3} & 57.09 & 51.47 & 53.78 \\
		LLaVA-3D~\cite{zhu2025llava} & 36.81 & 50.49 & 44.88 \\
		
		\midrule
		\multicolumn{4}{l}{\textbf{\textit{Video LLMs}}} \\
		LLaVA-Video~\cite{zhang2024llava} & 35.76 & 41.03 & 38.86 \\
		LLaVA-NeXT-Video & \underline{72.45} & 54.73 & \underline{62.00} \\
		InternVL-3.5~\cite{wang2025internvl3} & 63.54 & \underline{59.62} & 61.23 \\
		Qwen3-VL~\cite{bai2025qwen3} & 66.71 & 57.50 & 61.28 \\
		\rowcolor{gray!10} \textbf{AmbiVer} & \textbf{74.56} & \textbf{69.41} & \textbf{71.52} \\
		\bottomrule
	\end{tabular}
\end{table}

\subsection{Cross-Dataset Generalization}
\label{sec:generalization}

To rigorously evaluate the out-of-distribution (OOD) generalization capabilities of the models, we construct a novel evaluation set based on the challenging Mip-NeRF 360 dataset~\cite{barron2022mip}. Crucially, to ensure this OOD evaluation aligns with the high fidelity of our primary benchmark, we subjected these new scenes to the identical multi-stage, human-in-the-loop annotation pipeline detailed in Section~\ref{sec:dataset} (incorporating the strict unanimous agreement protocol for binary labels). This rigorous process generates 2,079 high-quality consensus-backed instructions anchored in 7 diverse unbounded scenes, including 3 outdoor scenes (bicycle, garden, stump) and 4 indoor scenes (bonsai, counter, kitchen, room).

The zero-shot cross-domain generalization results in Table~\ref{tab:mipnerf_results} reveal that existing end-to-end models exhibit severe performance instability across varying scenes.
For instance, while Video-3D LLM achieves 72.22\% accuracy in outdoor scenes, its performance drops significantly to 57.18\% indoors. Conversely, models like LLaVA-3D fail severely in outdoor settings, yielding an accuracy of merely 36.81\%. 
Overall, the peak average accuracy across all baselines is capped at 64.12\%. 
This limitation arises precisely because end-to-end models fail to effectively capture fine-grained semantic evidence when exposed to novel environments. 
In contrast, AmbiVer achieves an impressive average accuracy of 71.52\%, consistently outperforming all baselines across all scenarios. Such robust generalization compellingly demonstrates that our model can precisely capture fine-grained visual semantics even in complex, unseen environments, thereby ensuring highly reliable ambiguity detection.

\begin{table}[t]
	\centering
	\caption{Ablation study of the perception engine.}
	\label{tab:ablation_perception}
	\setlength{\tabcolsep}{4pt}
	\begin{tabular}{clcc}
		\toprule
		\textbf{Case} & \textbf{Method} & \textbf{Acc. $\uparrow$} & \textbf{Macro-F1 $\uparrow$} \\
		\midrule
		\#1 & w/o Instruction Decoupling & 62.06 & 52.42 \\
		\#2 & w/o Adaptive Keyframe Selection  & 62.04 & 63.37 \\
		\#3 & w/o 3D Fusion & 58.59 & 53.61 \\
		\#4 & w/o Refinement Weights & \underline{64.93} & \underline{65.03} \\
		\#5 & Full Perception & \textbf{66.16} & \textbf{66.15} \\
		\bottomrule
	\end{tabular}
\end{table}

\subsection{Ablation Study}
\label{sec:abs}

\noindent\textbf{Ablation Study on the Perception Engine}
To evaluate the contribution of the perception engine, we design four ablation variants by modifying key components of the pipeline:
\textbf{1)}~Case \#1 bypasses instruction preprocessing, directly feeding the raw instruction $T$ to the grounding model without extracting the specific target query $Q_t$.
\textbf{2)}~Case \#2 replaces adaptive keyframe selection with a uniform temporal sampling of $N_t$ frames.
\textbf{3)}~Case \#3 removes ray-based 3D geometric fusion. The top-$K$ 2D detections $\mathcal{D}$ are simply selected by confidence and used as local evidence $\mathcal{C}$ without filtering spatial redundancy.
\textbf{4)}~Case \#4 keeps 3D fusion but simplifies the refinement step, choosing the representative view $d_k^*$ solely based on detection confidence, thereby ignoring visibility scores ($w_{\text{vis}}$) and boundary penalties ($w_{\text{bnd}}$).
\textbf{5)}~Case \#5 represents our full pipeline.

The results are presented in Table~\ref{tab:ablation_perception}. 
Case \#1 shows a notable performance drop because inputting the instruction $T$ introduces linguistic noise. Consequently, the open-vocabulary detector fails to localize the core subject, resulting in irrelevant candidate extraction and subsequent VLM misjudgments. 
Case \#2 degrades performance because uniform sampling often misses rapid motions or crucial viewpoint changes, failing to capture the viewpoints required to resolve spatial occlusions. 
Notably, Case \#3 suffers the most severe degradation in accuracy, decreasing to 58.59\%. Without 3D fusion, multi-view 2D detections of the same object are mistakenly treated as separate instances. This geometric redundancy misleads the VLM into hallucinating multi-instance ambiguities. 
Case \#4 also exhibits a performance decline. Relying solely on detection confidence often yields occluded, truncated, or blurry crops near image boundaries, severely limiting the ability of the VLM to verify fine-grained attributes. 
Finally, Case \#5 demonstrates that each component is indispensable for constructing clean, geometrically consistent, and query-relevant 3D evidence.

\noindent\textbf{Ablation Study on the Reasoning Engine} 
To evaluate the necessity of each modality in the reasoning pipeline, we ablate different input components within the \texttt{Dossier} presented to the VLM:
\textbf{1)}~Case \#1 removes the global spatial context (the BEV map $\mathcal{I}_{\text{bev}}$).
\textbf{2)}~Case \#2 removes all local instance evidence, retaining only the BEV map and language input. 
\textbf{3)}~Case \#3 removes all visual information ($\mathcal{I}_{\text{bev}}$ and $\mathcal{C}$), providing only the raw instruction $T$ and text prompts. 
\textbf{4)}~Case \#4 is the full baseline utilizing all inputs.

The results are detailed in Table~\ref{tab:ablation_reasoning}. 
Case \#1 shows a sharp drop in Macro-F1 to 45.23\%. Without the BEV map, the VLM lacks the global spatial topology of the scene. Consequently, it struggles to confirm instance uniqueness or assess global spatial relations, leading to a substantial increase in false positives by misclassifying unambiguous spatial instructions. 
Case \#2 also experiences a significant performance decline. Without fine-grained local crops, the VLM fails to capture subtle visual details, making it unable to resolve attribute-level or state-level ambiguities, such as verifying whether a specific drawer is open or closed. 
Case \#3 yields the lowest accuracy. This confirms that relying solely on language priors without visual grounding leads to near-random guessing, highlighting the strong language bias of large language models. 
Finally, Case \#4 achieves the best performance, demonstrating that resolving complex 3D ambiguities requires integrating the global scene layout with detailed local visual features.

\begin{table}[t]
	\centering
	\caption{Ablation on the reasoning engine.}
	\label{tab:ablation_reasoning}
	\setlength{\tabcolsep}{8pt}
	\begin{tabular}{clcc}
		\toprule
		\textbf{Case} & \textbf{Method} & \textbf{Acc. $\uparrow$} & \textbf{Macro-F1 $\uparrow$} \\
		\midrule
		\#1 & w/o Global Context & 56.95 & 45.23 \\
		\#2 & w/o Local Evidence & \underline{60.71} & \underline{59.76} \\
		\#3 & w/o Visual Information & 53.99 & 53.78 \\
		\#4 & Full Reasoning & \textbf{66.16} & \textbf{66.15} \\
		\bottomrule
	\end{tabular}
\end{table}

\section{Conclusion}
In this paper, we introduce 3D Instruction Ambiguity Detection, a novel task crucial for reliable human-robot interaction in complex physical environments. To systematically evaluate this capability, we present Ambi3D, the first large-scale benchmark dedicated to identifying ambiguous language commands in 3D scenes. To tackle this challenge, we propose AmbiVer, a decoupled two-stage framework that effectively addresses the inherent limitations of existing end-to-end models on this ambiguity detection task. Experiments on Ambi3D demonstrate the value of our proposed task and benchmark, providing a solid foundation for verifiable embodied AI.

\noindent\textbf{Limitations \& Future Work.} First, upstream 3D perception bottlenecks our framework, as reasoning cannot recover missing evidence for undetected objects. Second, multistage inference latency potentially hinders practical real-time embodied deployment. Future work includes accelerating inference pipelines and extending this benchmark to dynamic environments.

	\bibliographystyle{ACM-Reference-Format}
	\bibliography{ref} 
	\clearpage   
	\appendix

\section{Construction and Analysis of Ambi3D}
This section details the complete pipeline for the Ambi3D benchmark, covering its construction and analysis. We first describe the data acquisition process for four distinct instruction types, followed by rigorous multi-stage annotation and quality control. Finally, we present a comprehensive statistical analysis of the resulting benchmark's properties.
\subsection{Data Acquisition}
\label{sec:dataset_acquisition}
Our data acquisition pipeline is designed to collect four distinct categories of instructions, which form the basis of our benchmark. 

\noindent\textbf{Grounded Instructions.} To acquire instructions grounded in 3D-centric human interactions, we bootstrap our dataset from ScanQA, leveraging its high-quality, human-annotated question-answer (QA) pairs as a semantic foundation. We posit that these QA pairs, being products of human annotation, offer a rich source of natural language reflecting genuine human queries about 3D environments. 
As illustrated in Figure~\ref{fig:grounded_example}, we employ a targeted prompting strategy to guide GPT-4o to automatically reformulate the question component of each QA pair into a semantically equivalent, executable instruction.
This methodology ensures our resulting instructions inherit the desirable properties of the source data, such as moderate length and rich semantic content, without being artificially simplistic.

\noindent\textbf{Unambiguous Instructions (Hard Negatives).}
To rigorously evaluate model robustness against superficial heuristics, we curated a challenging subset of instructions, termed ``Hard Negative Instructions''. These instructions are designed to appear superficially ambiguous, often referring to an object category with multiple instances (e.g., ``the chair''). However, they contain a unique descriptive phrase or qualifier (e.g., ``the chair by the window'') that precisely disambiguates the reference to a single object instance. Given the linguistic subtlety required for such valid yet challenging samples, this subset was meticulously authored by human experts to ensure accuracy and naturalness.

\noindent\textbf{Referential Ambiguity Instructions.}
We designed an automated pipeline to generate instructions with referential ambiguity (i.e., a non-unique target object) by leveraging ScanNet's object metadata and GPT-4o. This process systematically covers the three primary types of referential uncertainty: instance, attribute, and spatial ambiguity.
For \textbf{instance ambiguity}, the pipeline first identifies object classes with multiple instances in a scene (e.g., ``cup''). It then uses a specific prompt (Figure~\ref{fig:inst_ambi_example}) to generate instructions using only this class label, explicitly forbidding any distinguishing features.
For \textbf{attribute ambiguity}, using a different prompt (Figure~\ref{fig:attr_ambi_example}), the LLM is guided to apply subjective or relative adjectives (e.g., ``larger'') to multi-instance object classes, while being prohibited from using superlatives or other disambiguating qualifiers.
For \textbf{spatial ambiguity}, the process identifies common spatial pairs (e.g., ``chair'' and ``table'') and then prompts the LLM (Figure~\ref{fig:spat_ambi_example}) to use observer-dependent spatial terms (e.g., ``to the left of the table'') to construct the instruction.

\noindent\textbf{Action Ambiguity Instructions.}
In contrast to referential ambiguity, action ambiguity instructions employ a core action verb with multiple plausible and mutually exclusive interpretations. We employ an LLM-based approach for this. The pipeline first filters for objects suitable for diverse operations. It then utilizes a specific prompt (Figure~\ref{fig:act_ambi_example}) to guide GPT-4o to generate an instruction targeting the specific object ID but using an ambiguous action verb (e.g., ``handle'', ``adjust''), while ensuring that verbs with unambiguous intent are avoided.

\subsection{Data Annotation and Quality Control}

\label{sec:dataset_annotation}

\subsubsection{Annotator Information}
The dataset annotation was performed independently by 12 trained annotators, all possessing domain expertise in 3D scene understanding. To ensure high-quality labeling, all annotators were provided with a detailed tutorial and a comprehensive annotation manual, which included precise definitions for each ambiguity type and guidelines for handling boundary cases. Furthermore, each annotator was required to pass a qualification test to ensure full comprehension of the task standards before beginning.

\subsubsection{Annotation Process}
We designed a multi-stage annotation process to maximize label accuracy and consistency.

\noindent\textbf{Stage 1: Initial Data Cleaning and Filtering.} Before the primary annotation, we performed a rigorous cleaning pass. This stage first employed an automated script, based on exact matching, to identify and remove instructions that were duplicates within the same scene. Subsequently, human reviewers conducted a secondary audit to filter out other invalid instructions, including those that were \textbf{1)} grammatically incorrect or semantically nonsensical, \textbf{2)} irrelevant to the 3D scene context, or \textbf{3)} overly simplistic and lacking meaningful interaction. This two-step cleaning process ensured that instructions proceeding to the next stage were valid, unique, and held research value.

\noindent\textbf{Stage 2: Core Ambiguity Annotation.} 
Each valid instruction was independently assigned to three different annotators. Annotators first evaluated the scene context by visualizing the raw 3D point cloud data using MeshLab. Then, guided by the annotation manual, they provided a primary binary label (unambiguous or ambiguous). For instructions labeled \lq ambiguous\rq, they also selected the corresponding primary sub-type.

\noindent\textbf{Stage 3: Consistency Check and Final Selection.} We employed a strict unanimous agreement protocol for the final selection. To ensure maximum reliability for the primary task, we only retained instructions where all three annotators were in unanimous agreement on the binary label. Any sample with binary-level disagreement was discarded. For the retained ambiguous instructions, the final sub-type label was determined by a majority vote; instructions failing to achieve a majority were similarly discarded. This approach does not introduce errors, as different ambiguity types can co-exist (e.g., \enquote{Please handle the larger chair to the left of the table}, which can simultaneously satisfy spatial, attribute, and action ambiguity).

This rigorous multi-stage protocol yielded the final Ambi3D dataset. This high standard for label consistency ensures the reliability and precision of the benchmark, providing a high-quality supervisory signal for model training and evaluation.

\begin{figure}[t]
	\centering
	\includegraphics[width=\columnwidth]{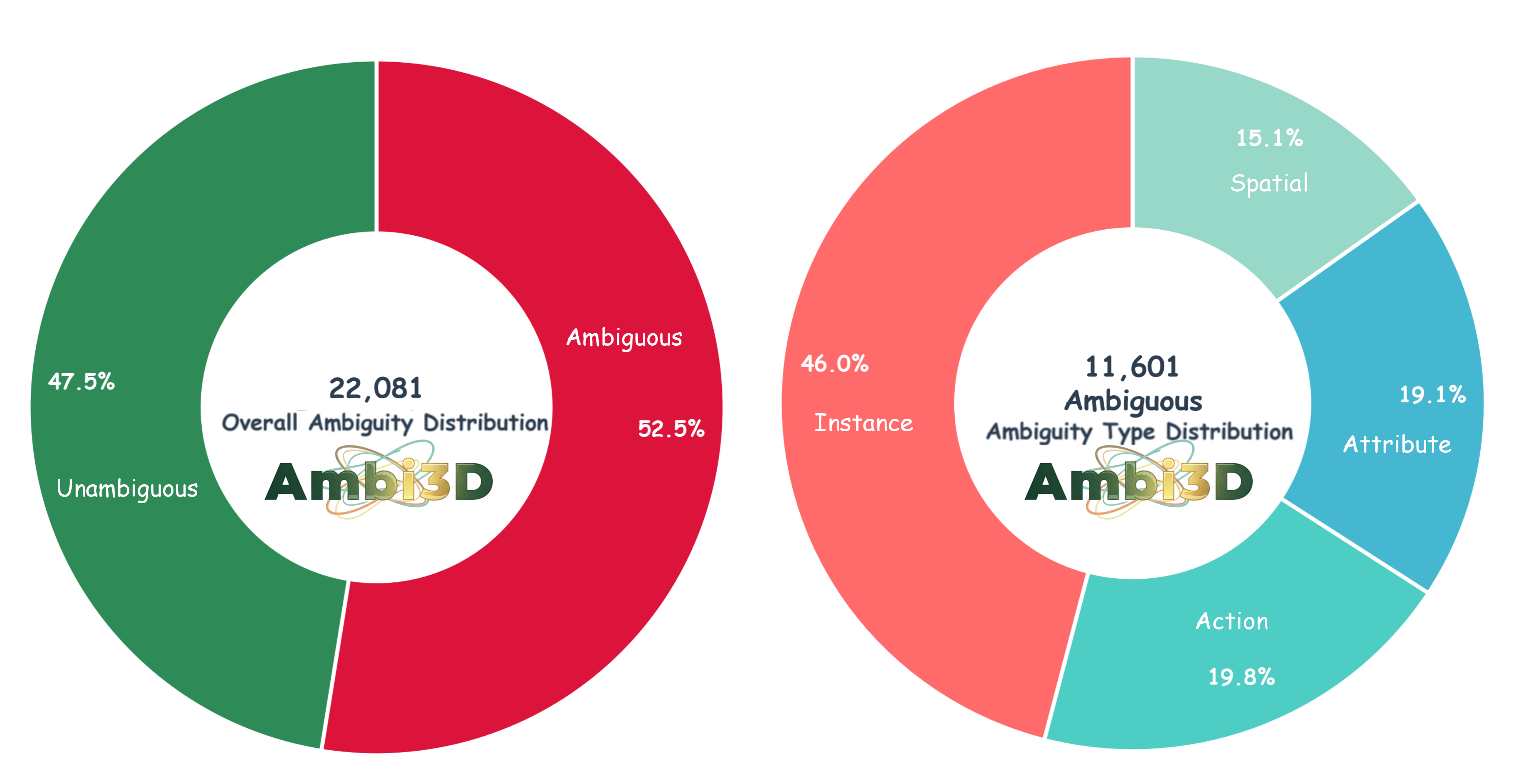}
	\caption{Ambi3D dataset composition. The analysis shows the overall class balance and the distribution of the four main ambiguity sub-types.}
	\label{fig:fig1}
\end{figure}

\subsection{Benchmark Analysis and Properties}
\noindent\textbf{Benchmark Composition.} The Ambi3D benchmark consists of 3D-grounded instructions categorized as either unambiguous or ambiguous. Figure~\ref{eui} provides qualitative examples of instructions from our dataset deemed unambiguous by the annotation process. Figure~\ref{eai} illustrates examples of the four primary ambiguity types annotated in our dataset: instance, attribute, spatial, and action. We now analyze the benchmark's statistical properties.

\noindent\textbf{Balance and Diversity}
An effective classification benchmark should avoid severe class imbalance. As shown in Figure~\ref{fig:fig1} (left), Ambi3D comprises 11,601 (52.5\%) ambiguous instructions and 10,480 (47.5\%) unambiguous instructions. This near 1:1 balanced distribution is crucial for preventing models from over-relying on the majority class prior during training. Furthermore, to ensure the task's comprehensiveness, we annotated multiple types of ambiguity. Figure~\ref{fig:fig1} (right) illustrates the internal composition of ambiguous instructions: instance ambiguity is the most prevalent (46.0\% of ambiguous samples), followed by action (19.8\%), attribute (19.1\%), and spatial (15.1\%) ambiguities. This diverse composition ensures that models must be capable of identifying different kinds of ambiguity, rather than achieving high scores merely by learning to resolve the most common type.

\begin{figure}[t]
	\centering
	\includegraphics[width=\columnwidth]{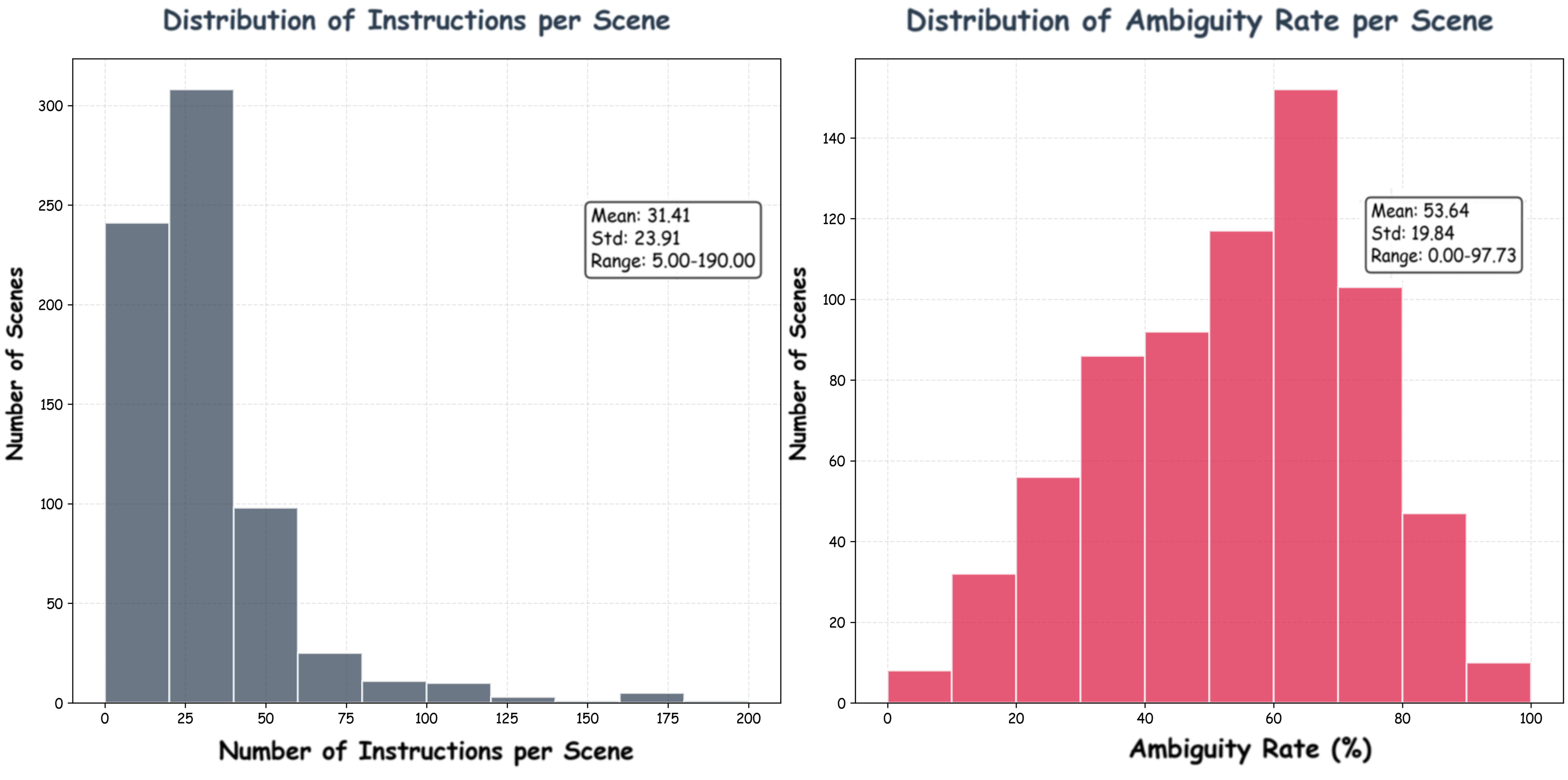}
	\caption{Ambi3D scene-level distributions. Histograms show the instruction counts per scene and the broad distribution of scene ambiguity rates.}
	\label{fig:fig2}
\end{figure}

\noindent\textbf{Avoiding Scene-Level Bias}
A critical risk in 3D datasets is that models may learn to exploit spurious correlations between global scene-level features and a target label. For example, if ambiguous instructions are highly concentrated in a few scenes, a model might learn a shortcut: directly associating the global visual representation of a scene with the ``ambiguous'' label, instead of performing the finer-grained contextual reasoning necessary to understand the instruction. To prevent this shortcut, we ensured that ambiguity is distributed broadly and heterogeneously across all 703 scenes. As shown in Figure~\ref{fig:fig2}, the ambiguity rate per scene exhibits a wide distribution, rather than being concentrated in a few scenes. This is a key design choice that compels the model to perform independent reasoning on each scene's context, rather than relying on this spurious, scene-level appearance-based correlation.

\begin{figure*}[t]
	\centering
	\includegraphics[width=\linewidth]{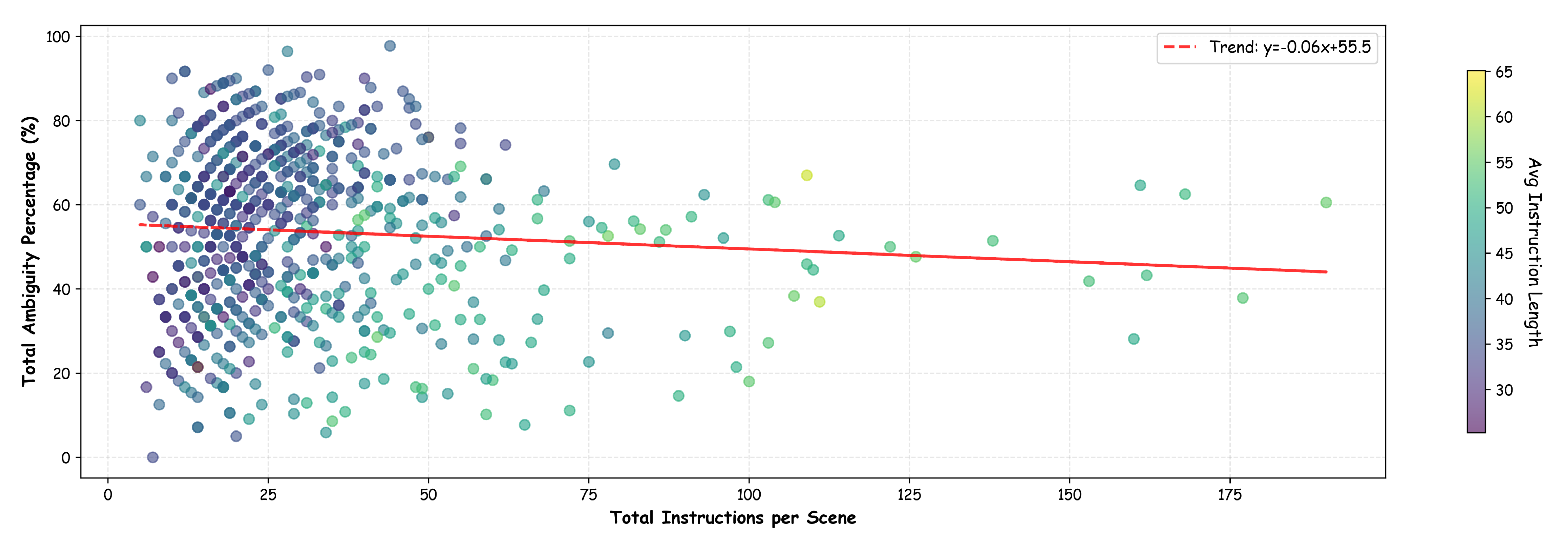}
	\caption{Analysis of advanced ambiguity patterns via heatmap and scatter plots.}
	\label{fig:fig4}
\end{figure*}

\begin{figure}[t]
	\centering
	\includegraphics[width=\columnwidth]{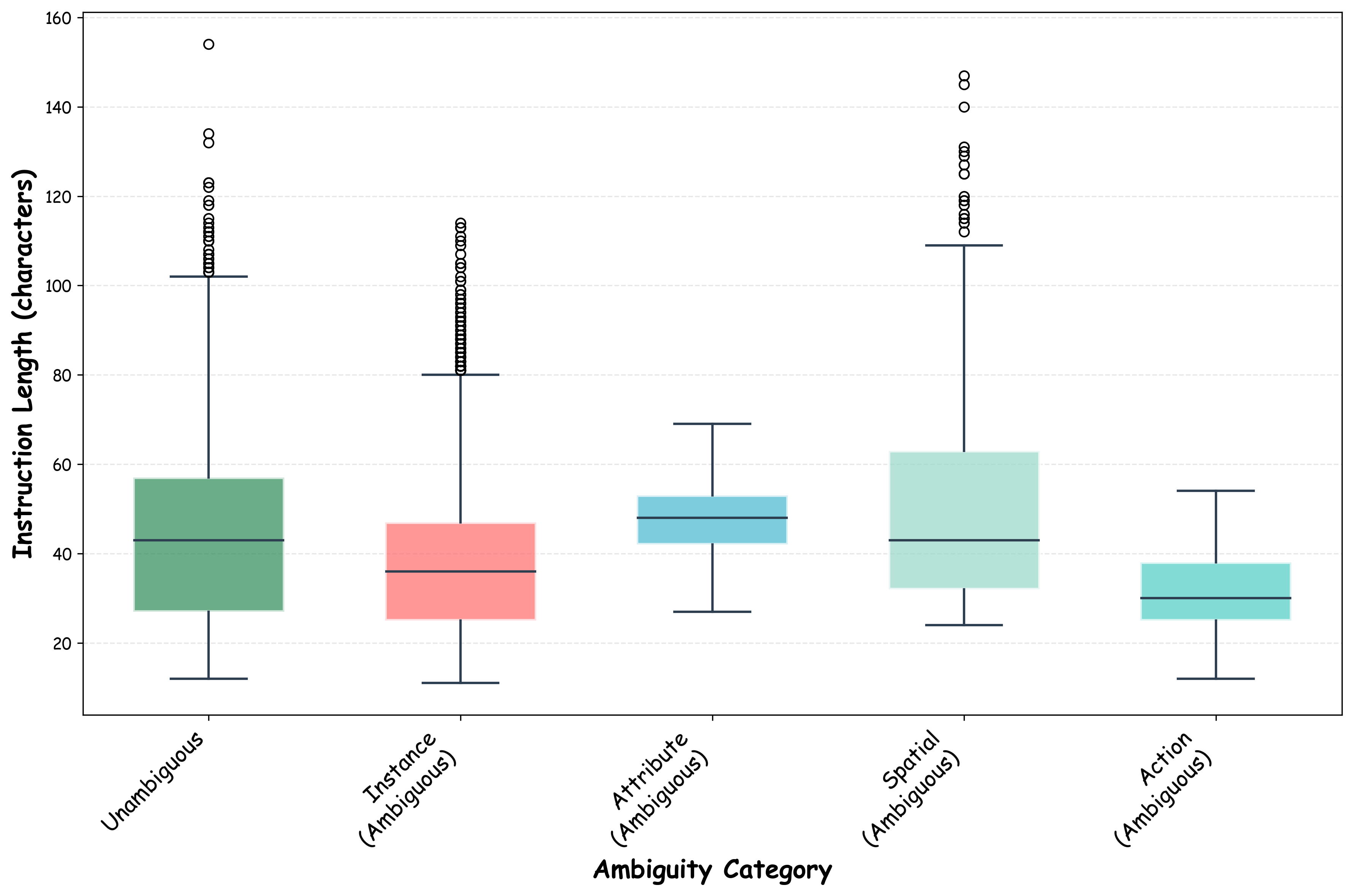}
	\caption{Instruction length distribution by ambiguity type. The significant overlap in distributions demonstrates that instruction length is not a reliable heuristic for distinguishing ambiguity.}
	\label{fig:fig3}
\end{figure}

\begin{figure}[t]
	\centering
	\includegraphics[width=\columnwidth]{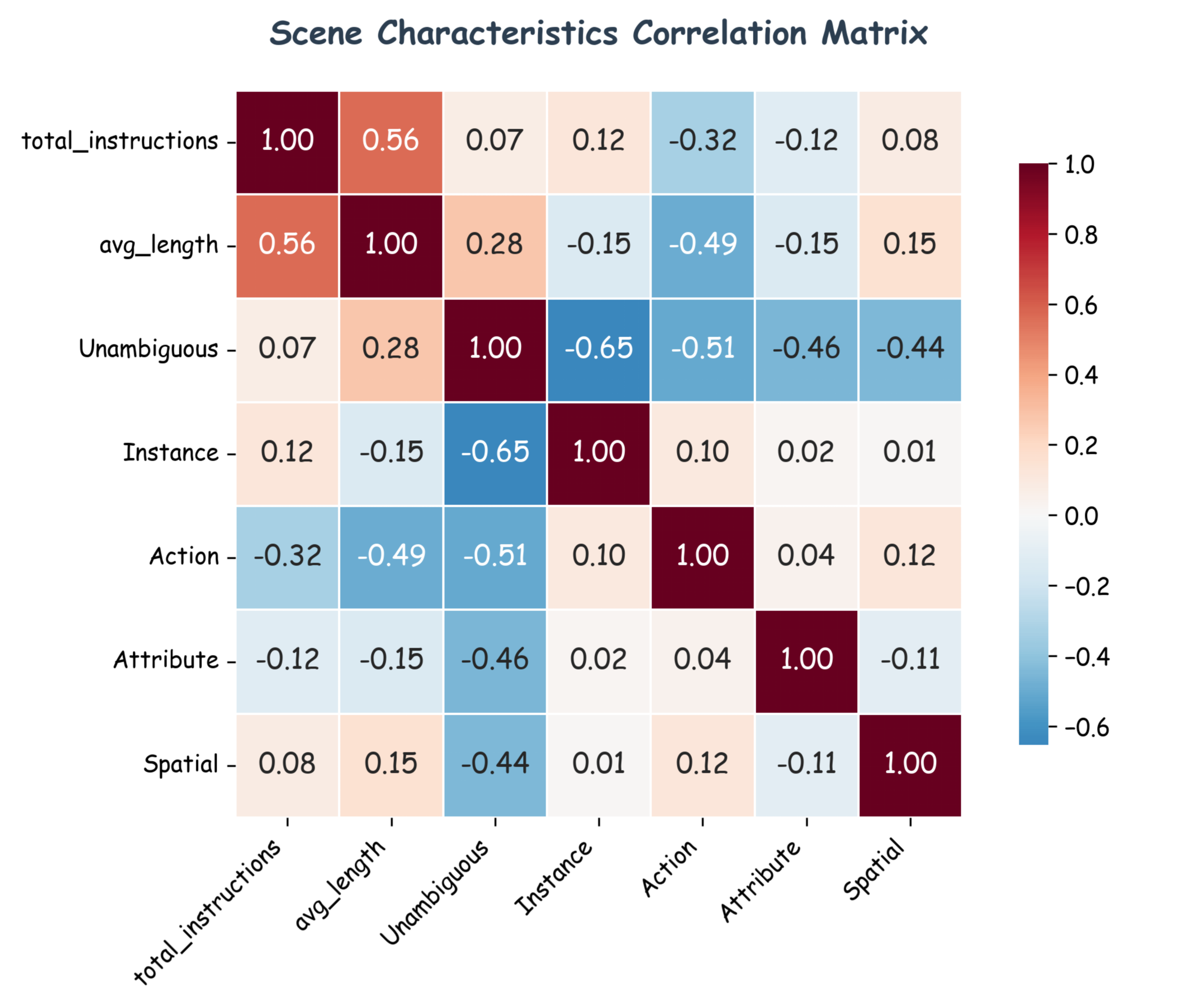}
	\caption{Analysis of scene size versus ambiguity rate. The scatter plot reveals no significant linear correlation between the total number of instructions per scene and the scene-level ambiguity rate, suggesting the task's non-linear complexity.}
	\label{fig:fig5}
\end{figure}

\noindent\textbf{Evading Surface Heuristics}
Simple models might attempt to learn spurious correlations between surface features and labels, such as ``shorter instructions are more likely to be ambiguous.'' We intentionally broke this potential association during data construction. As illustrated in Figure~\ref{fig:fig3}, the length distributions of ambiguous instructions highly overlap with that of unambiguous instructions, with very close means and medians. This design prevents models from distinguishing ambiguity using instruction length as a simple heuristic, forcing them to deeply understand the complex interaction between instruction semantics and their 3D visual context.

\noindent\textbf{Non-Linear Complexity}
Finally, we investigate deeper correlation patterns within the dataset to demonstrate the task's complexity. To conduct this scene-level analysis, we first iterate through all instructions and aggregate them by their scene ID. Through this process, we construct a unified 7-dimensional feature vector $V_s$ for each of the 703 scenes $s$:
\begin{equation}
	\label{eq:feature_vector}
	V_s = [I_s, L_s, P_s^{\text{unamb}}, P_s^{\text{inst}}, P_s^{\text{act}}, P_s^{\text{attr}}, P_s^{\text{spat}}]
\end{equation}
where $I_s$ is the total number of instructions in scene $s$ , $L_s$ is the average instruction length in scene $s$, and $P_s^{\text{unamb}} \dots P_s^{\text{spat}}$ are the percentages of unambiguous, instance, action, attribute, and spatial ambiguity instructions in scene $s$, respectively.

We first organize these 703 feature vectors $V_s$ into a $703 \times 7$ data table, where each column represents the values of a feature across all 703 scenes. We then compute the Pearson correlation coefficient between any two columns in this table.
As shown in Figure~\ref{fig:fig4}, the correlations between the percentages of different ambiguity types (e.g., $P^{\text{inst}}$ and $P^{\text{attr}}$) are generally weak, with coefficients near 0. This is an important design feature as it eliminates a potential statistical shortcut. If two ambiguity types were highly correlated (e.g., $r > 0.9$), a model might learn a spurious association. Our orthogonality implies that the presence of one ambiguity type in a scene has almost no predictive value for the presence of another, forcing the model to independently learn the specific linguistic and visual evidence for each type.

Furthermore, we analyze the relationship between scene size and its total ambiguity rate via a scatter plot. As depicted in Figure~\ref{fig:fig5}, the data points are widely scattered, and the linear regression trend line is nearly horizontal, strongly confirming the lack of a simple linear relationship between the two. The color of the points (representing $L_s$, average instruction length) also shows no obvious stratification. These findings jointly confirm that the ambiguity detection task proposed by Ambi3D is a complex, non-linear semantic reasoning challenge: its difficulty is not simply determined by the scene's physical complexity (e.g., number of objects) or the instruction's surface complexity (e.g., length), but by the fine-grained interaction between language and 3D context.

\begin{algorithm}[t]
	\caption{Instance Unification and Refinement Pipeline}
	\label{alg:perception_pipeline}
	\begin{algorithmic}[1]
		\Require Keyframes $\{I_v\}_{v=1}^{N_{\text{t}}}$, Poses $\{\mathcal{E}_v\}_{v=1}^{N_{\text{t}}}$, Query $Q_{\text{t}}$, Top $K$
		\Ensure Set of $K$ instance candidates $\mathcal{C}$
		
		\State $\mathcal{D} \gets \emptyset$
		\For{each keyframe $v \in [1, N_{\text{t}}]$}
		\State $\mathcal{D}_v \gets \Call{GroundingDINO}{I_v, Q_{\text{t}}}$
		\State $\mathcal{D} \gets \mathcal{D} \cup \{(v, b_i, s_i) \mid (b_i, s_i) \in \mathcal{D}_v\}$
		\EndFor
		
		\State $\mathcal{U} \gets \Call{InitializeUnionFind}{|\mathcal{D}|}$
		\For{each pair of detections $(d_i, d_j) \in \mathcal{D} \times \mathcal{D}$ where $v_i \neq v_j$}
		\State $\text{Ray}_i \gets \Call{BackProject}{d_i, \mathcal{E}_{v_i}}$
		\State $\text{Ray}_j \gets \Call{BackProject}{d_j, \mathcal{E}_{v_j}}$
		\If{$\Call{RayDist}{\text{Ray}_i, \text{Ray}_j} < \epsilon_d$}
		\If{$\Call{RayAngle}{\text{Ray}_i, \text{Ray}_j} \in [\theta_{a, \text{min}}, \theta_{a, \text{max}}]$}
		\If{$\Call{ScaleRatio}{d_i, d_j} > \sigma_s$}
		\State $\Call{Union}{\mathcal{U}, i, j}$
		\EndIf
		\EndIf
		\EndIf
		\EndFor
		\State $\{\mathcal{G}_k\} \gets \Call{GetConnectedComponents}{\mathcal{U}}$
		
		\State $\mathcal{C}_{\text{scored}} \gets \emptyset$
		\For{each group $\mathcal{G}_k \in \{\mathcal{G}_k\}$}
		\State $S_k \gets \Call{CalculateGroupScore}{\mathcal{G}_k}$
		\State $d_k^* \gets \Call{FindBestRepresentative}{\mathcal{G}_k}$
		\State $\mathcal{C}_k \gets (I_{v_k^*}, b_k^*, S_k, |\mathcal{G}_k|)$
		\State $\mathcal{C}_{\text{scored}} \gets \mathcal{C}_{\text{scored}} \cup \{\mathcal{C}_k\}$
		\EndFor
		
		\State $\mathcal{C}_{\text{ranked}} \gets \Call{RankByScore}{\mathcal{C}_{\text{scored}}}$
		\State $\mathcal{C} \gets \mathcal{C}_{\text{ranked}}[1...K]$
		\State \Return $\mathcal{C}$
	\end{algorithmic}
\end{algorithm}

\begin{algorithm}[t]
	\caption{Robust Label Extraction from LLM Response}
	\label{alg:extraction}
	\begin{algorithmic}[1]
		\Require Raw response text $T_{raw}$
		\Ensure Binary label $y \in \{0, 1\}$ (0: unambiguous, 1: ambiguous)
		
		\Function{ExtractLabel}{$T_{raw}$}
		\State $T \gets \Call{StripWhitespace}{T_{raw}}$
		
		\State $y_{num} \gets \Call{FindPriorityNumericMatch}{T}$
		\If{$y_{num}$ is not \textbf{null}}
		\State \Return $y_{num}$
		\EndIf
		
		\State $T_{lower} \gets \Call{ToLowerCase}{T}$
		
		\State $K_{unamb} \gets \{$ ``unambiguous'', ``not ambiguous'',
		\Statex \hspace{\algorithmicindent} ``clear'', ``specific'', ``precise'',
		\Statex \hspace{\algorithmicindent} ``definite'' $\}$
		\If{$\Call{ContainsAny}{T_{lower}, K_{unamb}}$}
		\State \Return 0
		\EndIf
		
		\State $K_{amb} \gets \{$ ``ambiguous'', ``unclear'', ``vague'',
		\Statex \hspace{\algorithmicindent} ``confusing'', ``uncertain'', ``multiple'' $\}$
		\If{$\Call{ContainsAny}{T_{lower}, K_{amb}}$}
		\State \Return 1
		\EndIf
		
		\State \Return 0
		\EndFunction
	\end{algorithmic}
\end{algorithm}

\section{Model Details}
\label{sec:appendix_impl_details}

This section provides detailed implementation specifics for the perception engine and reasoning engine.

\subsection{Perception Engine Details}

\noindent \textbf{Global Feature Acquisition.}
The BEV map $\mathcal{I}_{\text{bev}}$ is generated in two steps. First, the reconstruction pipeline $\mathcal{R}$ uses BundleFusion~\cite{dai2017bundlefusion} to aggregate the video $\mathcal{V}$ and poses $\mathcal{E}$ into a unified 3D point cloud $\mathcal{P}$. Second, the transformation $\mathcal{T}$ projects $\mathcal{P}$ into $\mathcal{I}_{\text{bev}}$ using a fixed top-down camera pose $\mathcal{E}_{\text{top}}$.

\noindent \textbf{Adaptive Keyframe Selection.}
Processing all frames is prohibitive due to the high computational cost. We implement the adaptive keyframe selection as follows: We always select the first frame. Then, we iterate through the remaining frames, calculating the pose dissimilarity between the current frame $t$ and the last selected keyframe $t_{\text{last}}$. Dissimilarity is measured by both translational distance $d_t = \| \mathbf{p}_t - \mathbf{p}_{t_{\text{last}}} \|_2$ and the maximum absolute Euler angle difference $d_r = \| \text{euler}(\mathbf{R}_t \mathbf{R}_{t_{\text{last}}}^T) \|_{\infty}$. A frame is selected if $d_t > \tau_t$ or $d_r > \tau_r$. We set the initial thresholds $\tau_t=0.15$m and $\tau_r=15.0^{\circ}$.
If the final keyframe count $N_c$ deviates significantly from the target $N_{\text{t}}$ (e.g., 100), the thresholds are adaptively adjusted and the process is re-run, as described in the main paper.

\noindent \textbf{Instance Unification and Refinement.}
The core of the \lq Detailed Feature Acquisition\rq is formalized in Algorithm~\ref{alg:perception_pipeline}. This process encompasses candidate localization, 3D ray-based geometric merging using a Union-Find data structure, and the final scoring and selection of the top-$K$ candidates.

\noindent \textbf{Representative Score Details.}
The selection of a representative detection $d_k^*$ is performed by the $\Call{FindBestRepresentative}{\mathcal{G}_k}$ function, which maximizes the composite score $f(d_i)$. As described in the main text, this score balances confidence $s_i$, visibility $w_{\text{vis}}$, and a boundary penalty $w_{\text{bnd}}$.
The visibility term $w_{\text{vis}}(d_i)$ is calculated as the bounding box area $b_i$ relative to its image $I_{v_i}$:
\begin{equation}
	w_{\text{vis}}(d_i) = \text{area}(b_i) / \text{area}(I_{v_i})
	\label{eq:appendix_vis}
\end{equation}
The boundary penalty $w_{\text{bnd}}(d_i)$ is set to $\gamma$ if the box is within $\delta$ pixels of any image border, and $1.0$ otherwise.

\noindent \textbf{Default Hyperparameters.} We employ Qwen-3-VL-8b-Instruct for the perception engine with the following default hyperparameters: $\tau_t=0.15$m, $\tau_r=15.0^{\circ}$, $\epsilon_d=0.3$m, $[\theta_{a, \text{min}}, \theta_{a, \text{max}}] = [0.0^{\circ}, 60.0^{\circ}]$, $\sigma_s=0.2$, $K=6$, $\gamma=0.5$, and $\delta=4$.

\subsection{Reasoning Engine Details}
\label{sec:appendix_prompt}
The multi-modal prompt is constructed from the $\texttt{Dossier}$, illustrated in Figure~\ref{fig:am_prompt_design}, and comprises two key components. First, a system message defines the execution-oriented ambiguity criterion for the VLM. Second, a user message presents the aggregated evidence, which contains both the global context (e.g., the BEV map) and all local instance-level evidence. This structured approach ensures that all global and local evidence is delivered to the VLM in a consistent and interpretable format for its final adjudication.

\section{Efficiency Analysis}
\label{sec:efficiency}

This section details the computational efficiency of AmbiVer. We evaluate the inference latency averaged across all instructions in the Ambi3D dataset using a single NVIDIA RTX 4090 GPU. 

As summarized in Table~\ref{tab:efficiency}, AmbiVer achieves an average overall latency of approximately 7.51 seconds. We decompose the system pipeline into two primary stages: perception and reasoning. The perception stage accounts for the majority of the computational cost, primarily bottlenecked by the visual grounding model (Grounding DINO), which takes 4.975 seconds. In contrast, the remaining perception components, including instruction parsing, adaptive keyframe selection, and ray-based union-find, are highly efficient and operate in under 0.1 seconds combined. The subsequent reasoning stage, which involves Vision-Language Model (VLM) adjudication, takes 2.450 seconds. 

Overall, this execution time is acceptable for static ambiguity resolution tasks in robotic navigation. In such scenarios, the agent typically pauses to analyze the environment and resolve linguistic uncertainties before initiating physical movement.

\begin{table}[t]
	\centering
	\caption{Detailed latency breakdown of AmbiVer averaged over the Ambi3D dataset.}
	\label{tab:efficiency}
	\begin{tabular}{llc}
		\toprule
		\textbf{Stage} & \textbf{Component} & \textbf{Time (s)} \\
		\midrule
		Perception & Instruction Parsing & 0.001 \\
		& Adaptive Keyframe Selection & 0.010 \\
		& Grounding DINO & 4.975 \\
		& Ray-based Union-Find & 0.075 \\
		\midrule
		Reasoning  & VLM Adjudication & 2.450 \\
		\midrule
		\textbf{Total} & \textbf{Overall Latency} & \textbf{7.511} \\
		\bottomrule
	\end{tabular}
\end{table}

\section{Analysis of Failure Cases} 
While AmbiVer significantly outperforms existing baselines, which often fail completely in ambiguous 3D scenarios, our method still exhibits certain limitations. We observe two primary types of remaining errors: missed ambiguities (false negatives) and over-sensitivity (false positives).

\noindent \textbf{Missed Ambiguities (False Negatives).} Although our model detects conflicts much better than prior works, it occasionally misses multi-instance conflicts (e.g., ``bring me the cup'' when multiple cups exist). This occurs because distinguishing identical instances from BEV and cropped visual representations remains inherently challenging. Similarly, the model sometimes overlooks action ambiguities, struggling to reason about verb polysemy and underspecified execution methods (e.g., ``adjust the window''). Spatial false negatives also occur when the model neglects observer-dependent relative positions, missing the perspective ambiguity in phrases like ``the rightmost cabinet.''

\noindent \textbf{Over-Sensitivity (False Positives).} Conversely, our model can be overly cautious. It occasionally flags clear commands with globally unique targets (e.g., ``turn on the smallest monitor'' when only one such monitor exists) or explicit actions (e.g., ``clean the floor'') as ambiguous. While baseline models typically fail by missing ambiguities entirely, our model's occasional over-sensitivity suggests that exhaustively confirming object uniqueness and clear semantics within complex 3D scenes remains an open challenge.

\section{Experiment Details}
\subsection{Baseline Model Adaptation}
\label{sec:baseline_adaptation}

\noindent\textbf{Prompt Design.}
To adapt pre-trained language models to our ambiguity detection task, we designed a structured prompt template.
This template, detailed in Figure~\ref{fig:prompt_design}, adopts a few-shot paradigm. It first explicitly defines the task objective and the five primary ambiguity types (e.g., object, action, spatial ambiguity). It then strictly constrains the output format, requiring the model to respond with only a single digit: ``0'' (unambiguous) or ``1'' (ambiguous).

\noindent\textbf{Inference and Prediction Extraction.}
During inference, the model generates responses autoregressively.
To reliably extract a binary classification result from the model's free-text output, which may occasionally contain formatting variations (e.g., extra text or whitespace), we implement the robust prediction extraction algorithm detailed in Algorithm~\ref{alg:extraction}.
This adaptation method leverages the pre-trained model's language understanding capabilities and successfully adapts the open-domain generative model to our required binary classification task through controlled output and reliable post-processing.

\subsection{LoRA Fine-tuning Details}
\label{sec:lora_details}
The prompt used during fine-tuning is identical to the one used at test time, with ``0'' and ``1'' serving as the ground-truth responses.
As detailed in Figure~\ref{fig:prompt_design}, this single prompt template is applied uniformly across all baseline models to ensure consistency.

\subsection{AmbiVer Prediction Handling}
\label{sec:am_prediction_handling}
To ensure a fair comparison, AmbiVer employs the same prompt template during evaluation, with the only modification being the required output format (detailed in Figure~\ref{fig:am_prompt_design}).
Unlike the free-text responses of the baselines, AmbiVer's output is consistently well-structured.
Consequently, the robust parsing logic from Algorithm~\ref{alg:extraction} is unnecessary. We instead directly extract the final prediction from the \texttt{label} field of the structured \texttt{verdict}.

\subsection{Details of the Mip-NeRF 360 Evaluation Set}
\label{sec:appendix_mipnerf}

To support the cross-dataset generalization experiments in the main text, this section provides additional details regarding the Mip-NeRF 360 evaluation set. Specifically, we clarify the handling of camera poses and visual evidence extraction in these unbounded scenes, and we report detailed dataset statistics.

For camera poses and visual evidence extraction, we directly use the standard camera trajectories provided by the Mip-NeRF 360 dataset. These trajectories typically consist of inward-facing views that capture the central region of the scene from 360 degrees. We render the scene frames using these provided poses. Following this, we apply the exact same evidence extraction pipeline used for our primary Ambi3D dataset. This unified approach ensures that the visual perception process remains consistent across datasets, and any performance variation is strictly due to out-of-distribution environments.

To ensure the navigation instructions follow the same distribution as Ambi3D, we used the identical annotation pipeline. This process yielded 2079 instructions across 7 scenes. The dataset contains 1337 unambiguous instructions and 742 ambiguous instructions. The ambiguous samples cover various types, including spatial ambiguity (639 cases), part-whole ambiguity (103 cases), and texture or instance ambiguity (3 cases). 

The instruction text length aligns with our primary benchmark, with an average length of 72.5 characters and a median of 75 characters. Table~\ref{tab:mipnerf_stats} details the number of instructions and the average character length for each individual scene.

\begin{table}[t]
	\centering
	\caption{Detailed statistics of the Mip-NeRF 360 evaluation set per scene.}
	\label{tab:mipnerf_stats}
	\begin{tabular}{lcc}
		\toprule
		\textbf{Scene} & \textbf{Total Instructions} & \textbf{Average Length} \\
		\midrule
		bicycle & 268 & 78.8 \\
		bonsai & 282 & 75.6 \\
		counter & 339 & 62.3 \\
		garden & 349 & 79.5 \\
		kitchen & 289 & 72.6 \\
		room & 316 & 67.6 \\
		stump & 236 & 72.4 \\
		\midrule
		\textbf{Total / Average} & 2079 & 72.5 \\
		\bottomrule
	\end{tabular}
\end{table}

\begin{figure*}[t]
	\centering
	\includegraphics[width=0.9\linewidth]{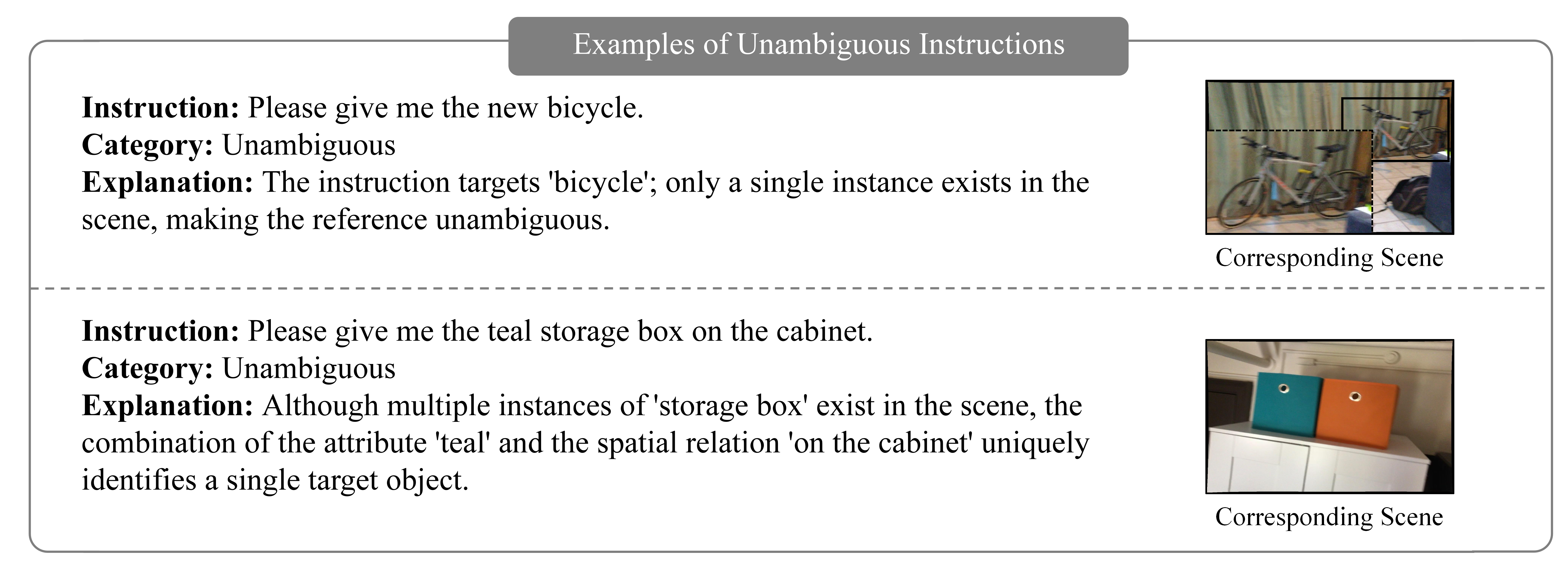}
	\caption{Qualitative examples of unambiguous instructions in Ambi3D. }
	\label{eui}
\end{figure*}

\begin{figure*}[t]
	\centering
	\includegraphics[width=0.9\linewidth]{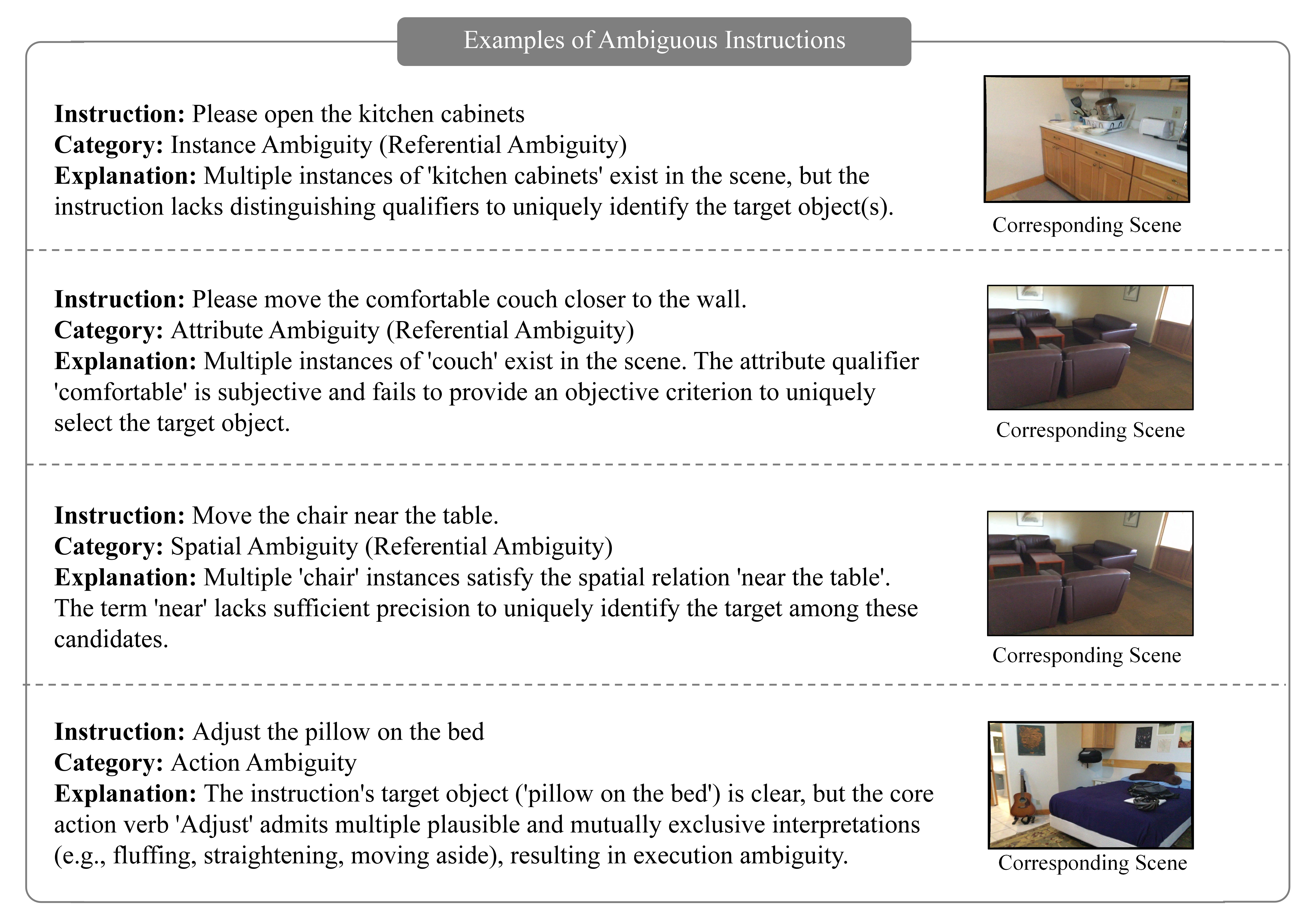}
	\caption{Qualitative examples of ambiguous instructions. }
	\label{eai}
\end{figure*}

\begin{figure*}[h!]
	\centering
	\includegraphics[width=0.9\linewidth]{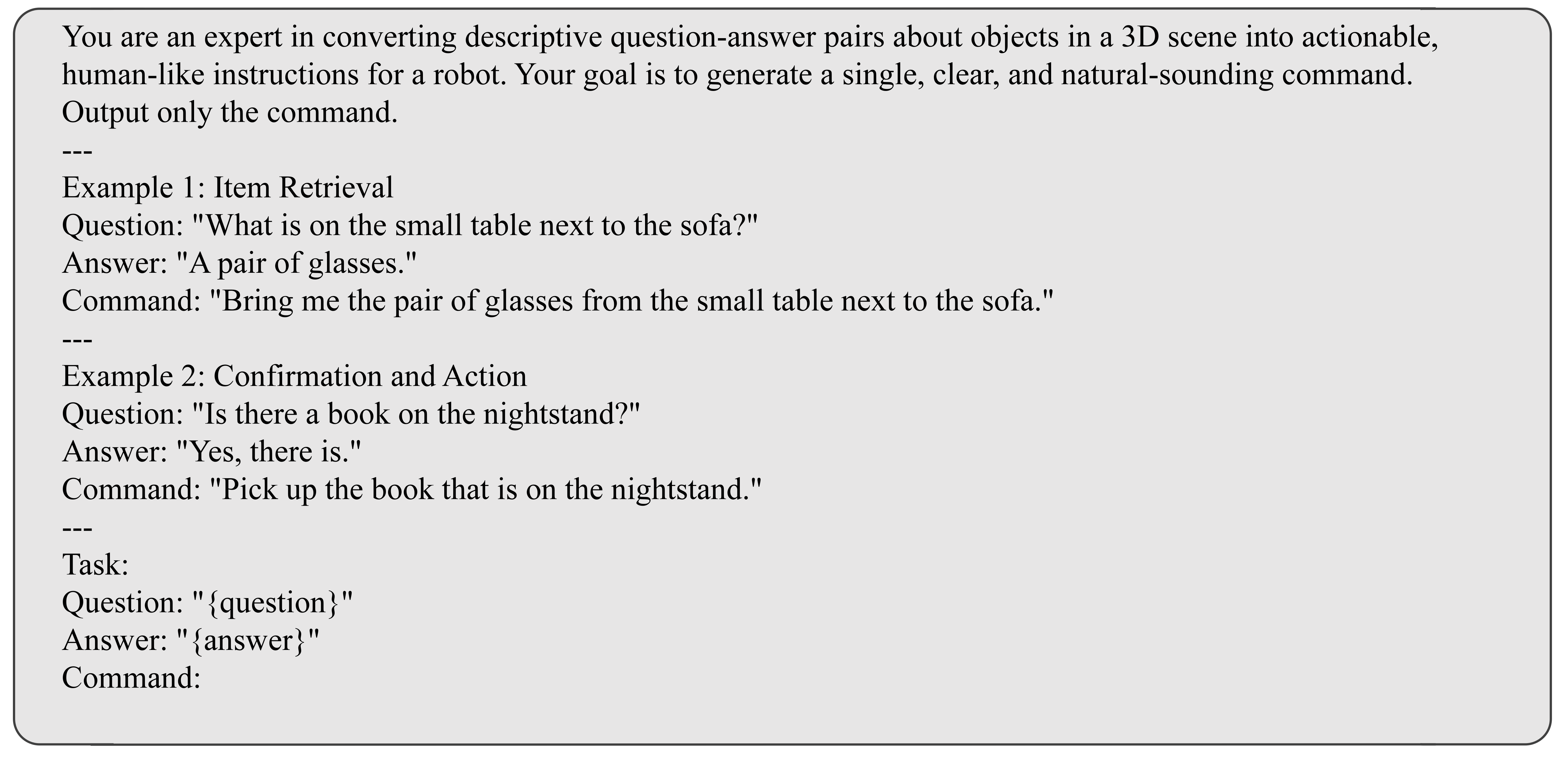}
	\caption{Prompting strategy to reformulate ScanQA questions into executable instructions.}
	\label{fig:grounded_example}
\end{figure*}

\begin{figure*}[h!]
	\centering
	\includegraphics[width=0.9\linewidth]{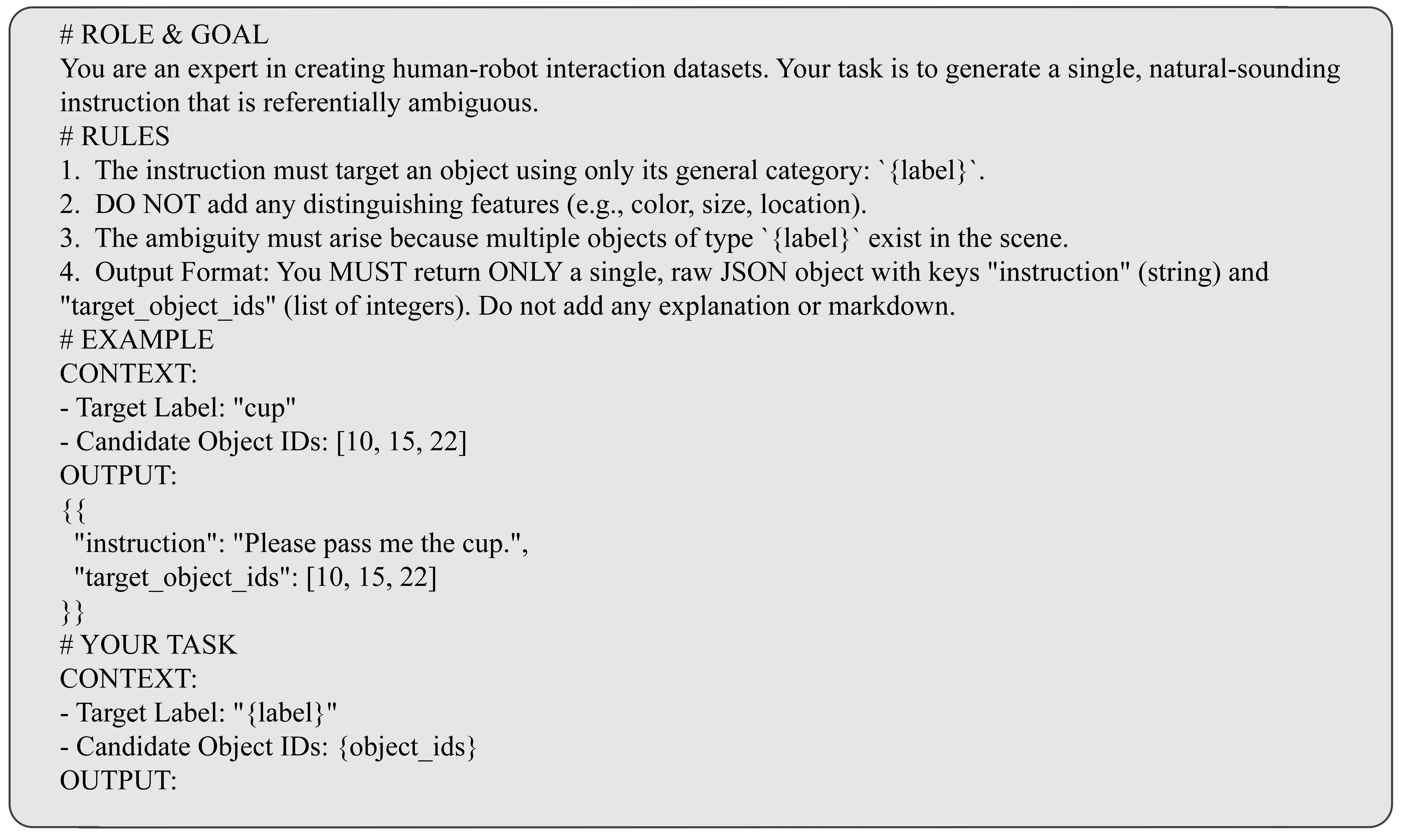}
	\caption{Prompting strategy for generating instructions with instance ambiguity.}
	\label{fig:inst_ambi_example}
\end{figure*}

\begin{figure*}[h!]
	\centering
	\includegraphics[width=0.9\linewidth]{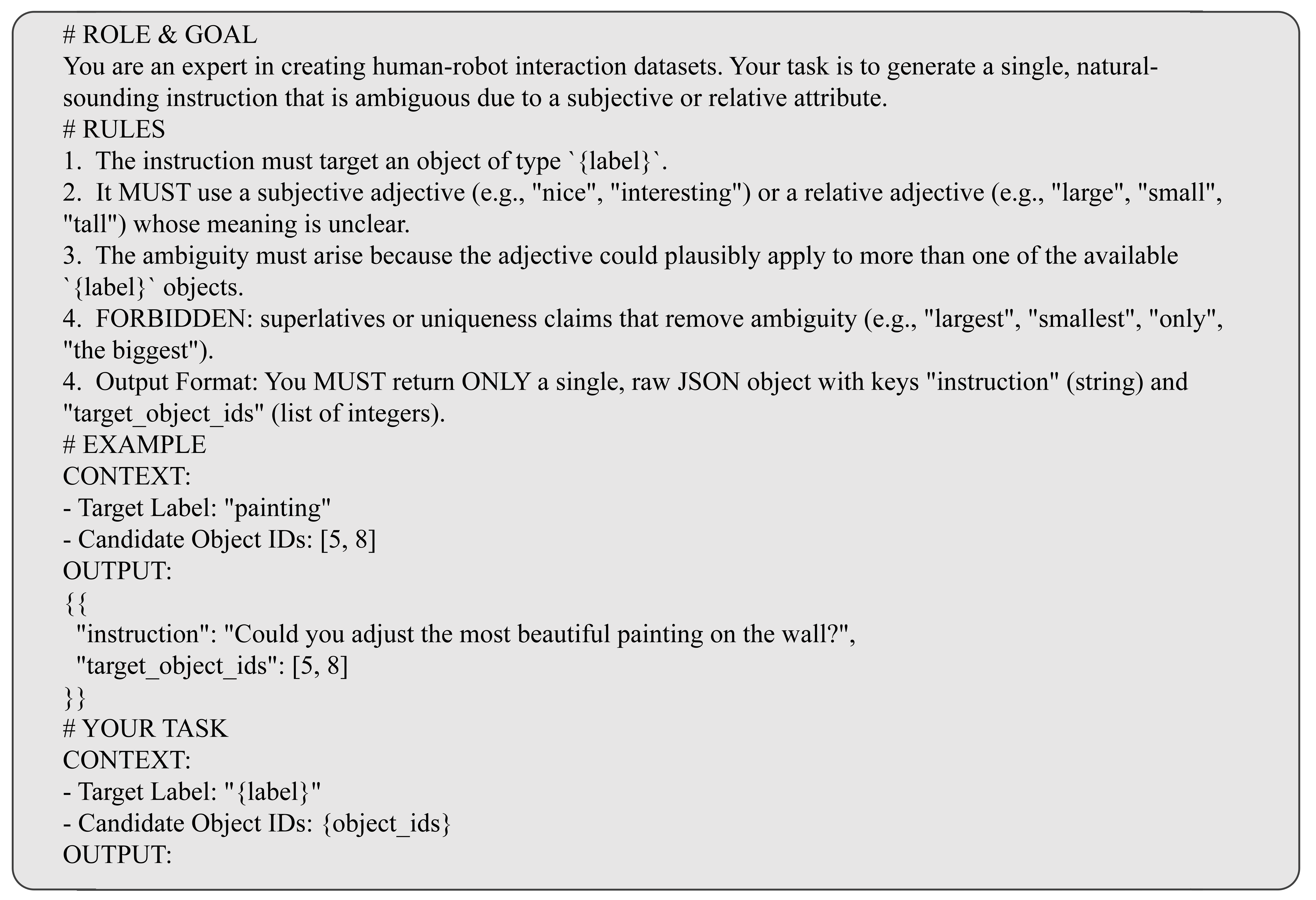}
	\caption{Prompting strategy for generating instructions with attribute ambiguity.}
	\label{fig:attr_ambi_example}
\end{figure*}

\begin{figure*}[h!]
	\centering
	\includegraphics[width=0.9\linewidth]{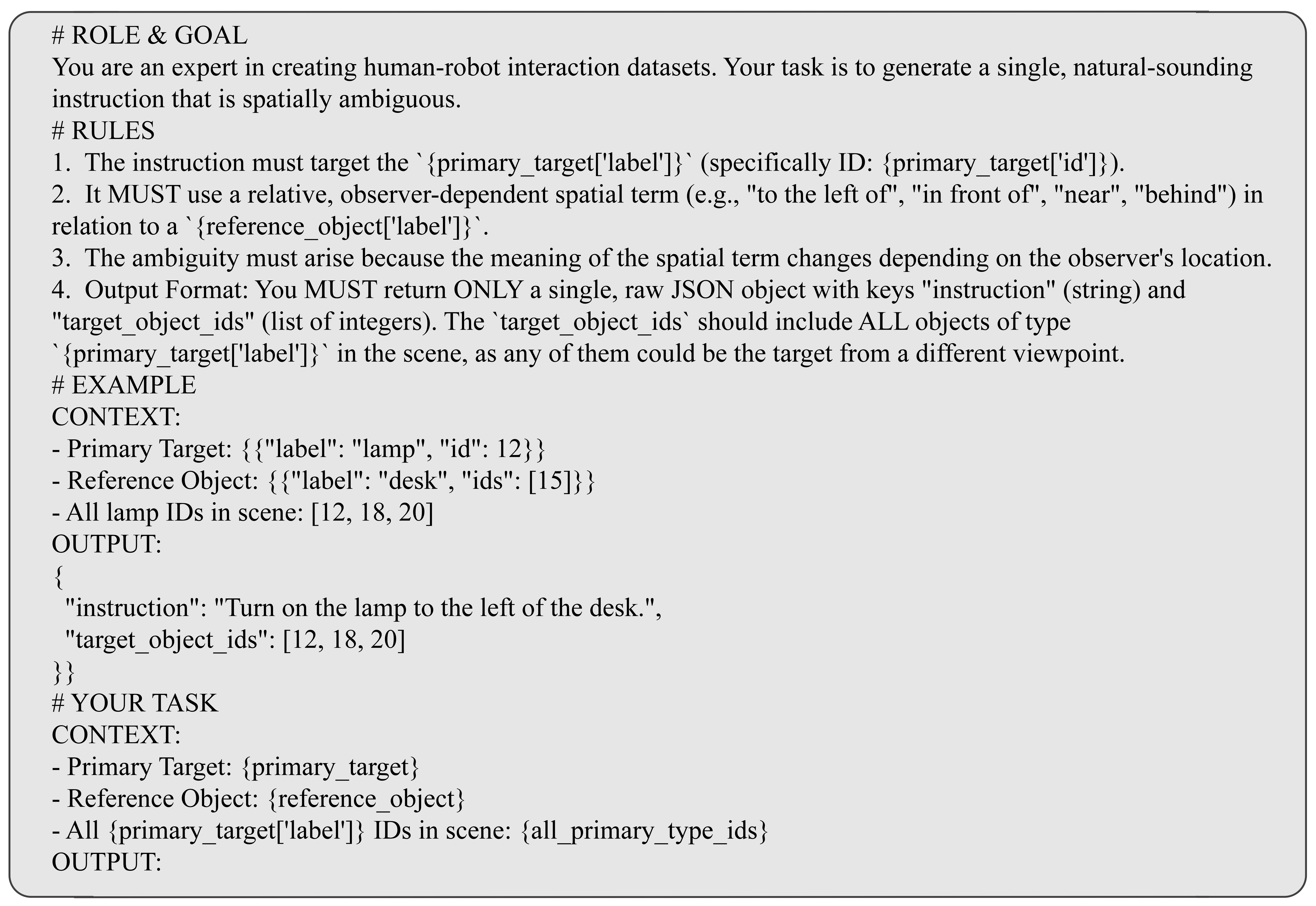}
	\caption{Prompting strategy for generating instructions with spatial ambiguity.}
	\label{fig:spat_ambi_example}
\end{figure*}

\begin{figure*}[h!]
	\centering
	\includegraphics[width=0.9\linewidth]{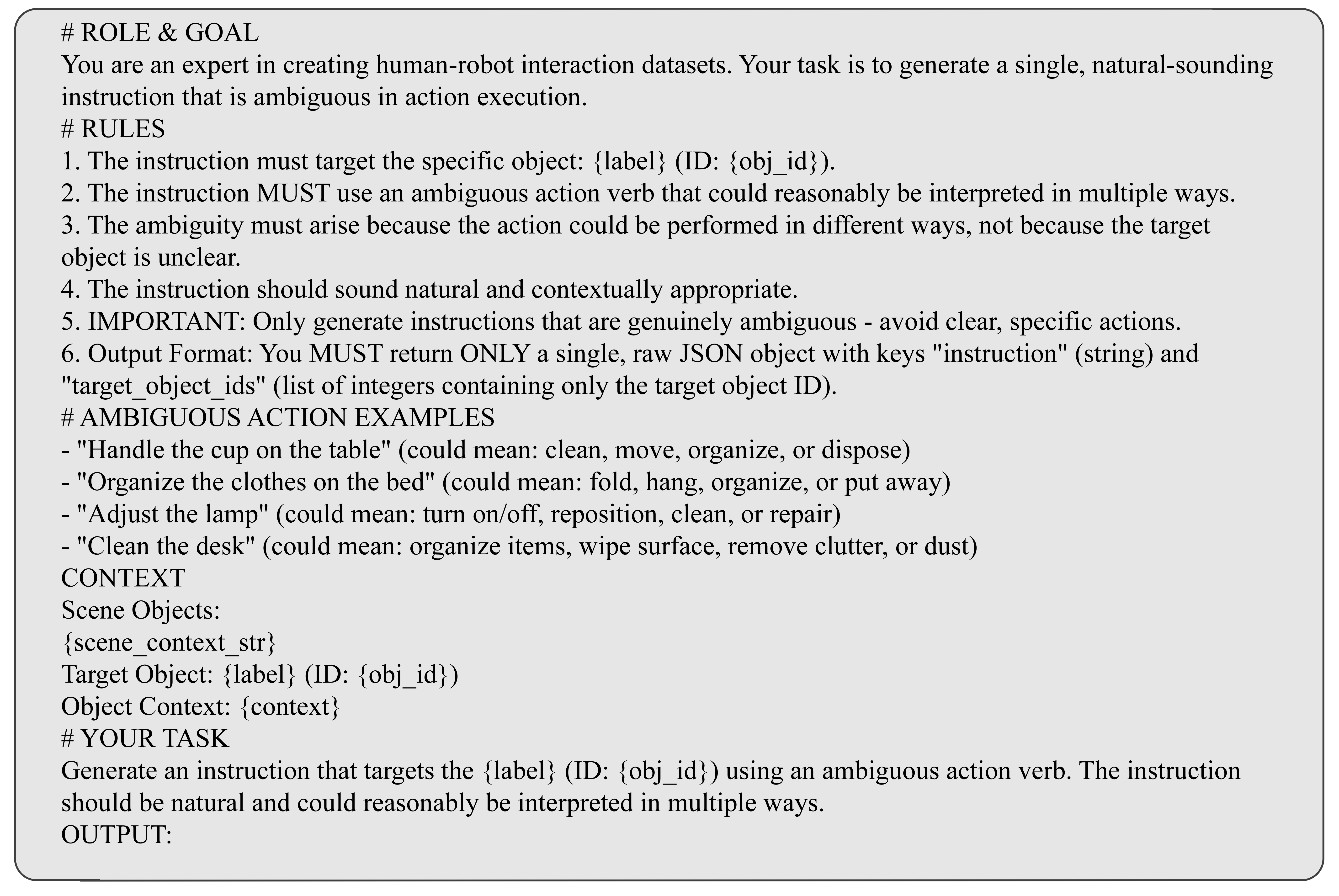}
	\caption{Prompting strategy for generating instructions with action ambiguity.}
	\label{fig:act_ambi_example}
\end{figure*}

\begin{figure*}[h!]
	\centering
	\includegraphics[width=0.9\linewidth]{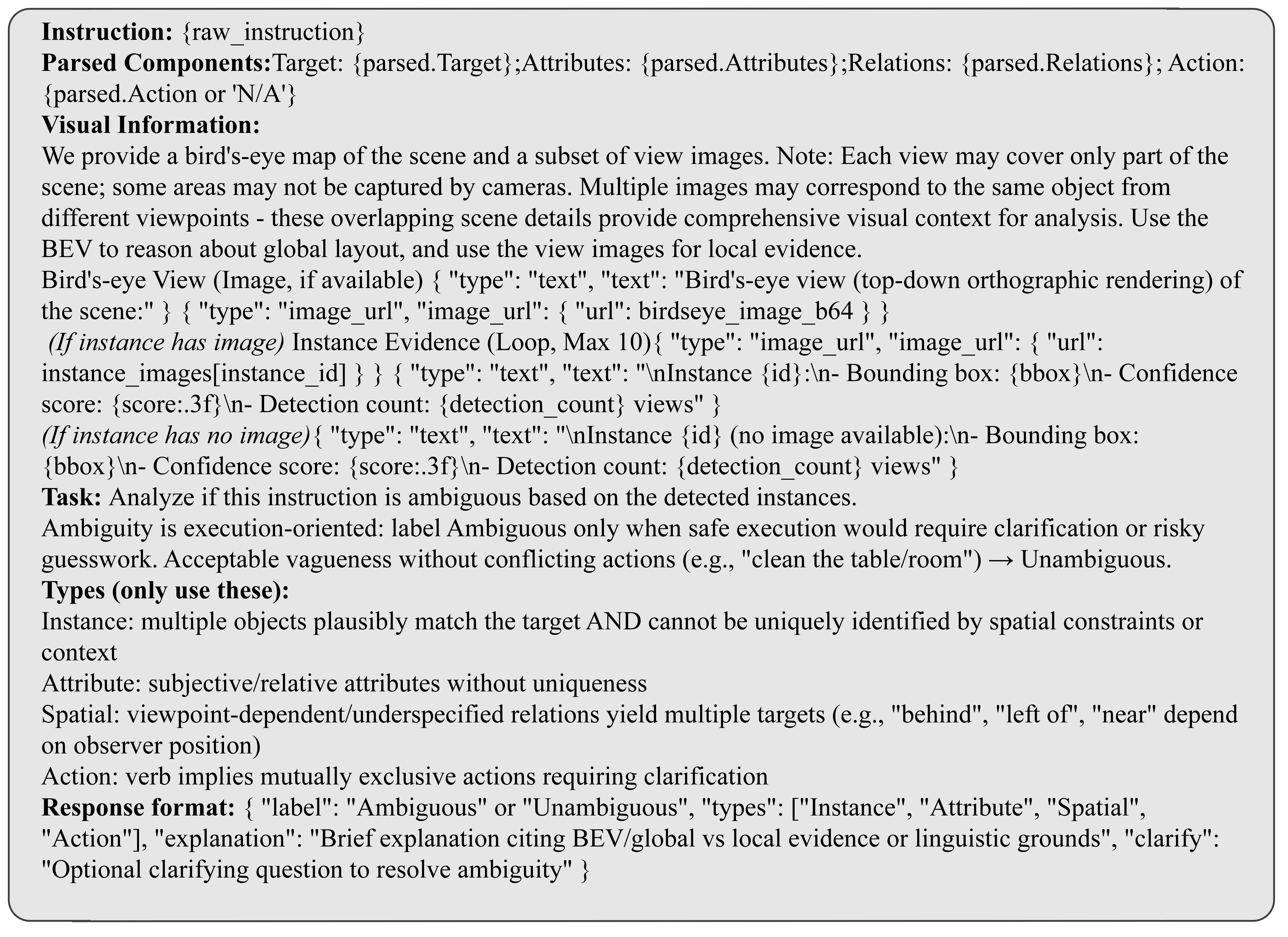}
	\caption{Prompting strategy for AmbiVer, which enforces a structured \texttt{verdict} output with a dedicated \texttt{label} field.}
	\label{fig:am_prompt_design}
\end{figure*}
\begin{figure*}[h!]
	\centering
	\includegraphics[width=0.9\linewidth]{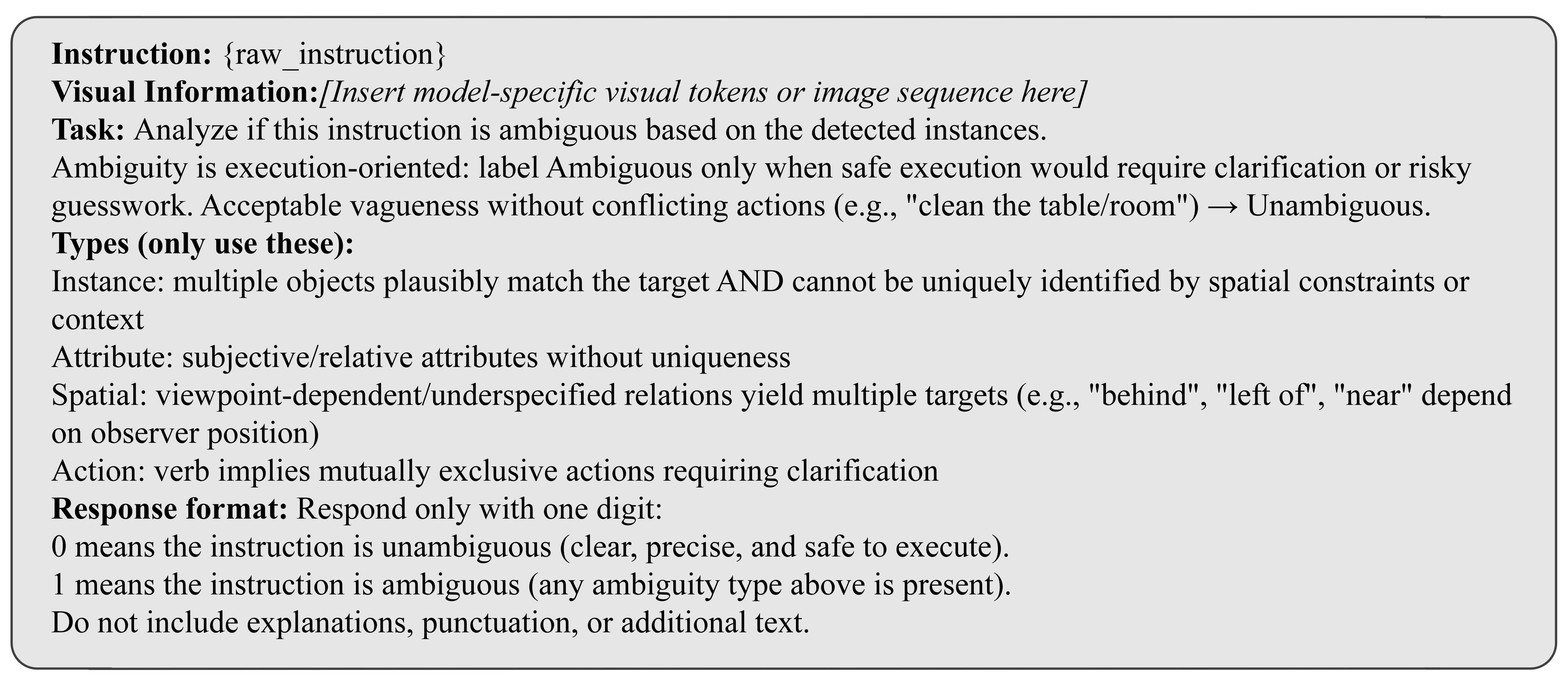}
	\caption{Prompting strategy for baseline model adaptation. This structured prompt defines the task, specifies ambiguity types, and constrains the output to a single binary digit.}
	\label{fig:prompt_design}
\end{figure*}

\end{document}